\documentclass[11pt]{article}
\usepackage[final]{acl}
\usepackage{times}
\usepackage{latexsym}
\usepackage[T1]{fontenc}
\usepackage[utf8]{inputenc}
\usepackage{microtype}
\usepackage{graphicx}
\usepackage{subcaption}
\usepackage{booktabs}
\usepackage{hyperref}
\usepackage{amsmath}
\usepackage{amssymb}
\usepackage{mathtools}
\usepackage{amsthm}
\usepackage{float}
\usepackage{algpseudocode}

\usepackage[capitalize,noabbrev]{cleveref}
\usepackage{comment}
\theoremstyle{plain}

\theoremstyle{definition}

\theoremstyle{remark}

\usepackage[textsize=tiny]{todonotes}
\usepackage{url}
\usepackage{enumitem}
\usepackage{multirow}
\usepackage{listings}
\usepackage{setspace}
\usepackage{placeins}
\usepackage[table]{xcolor}
\usepackage{inconsolata}
\usepackage{tcolorbox}
\usepackage{tabularx}
\usepackage{makecell}
\usepackage{algorithm}
\usepackage{colortbl}
\usepackage{pifont}
\tcbuselibrary{listings, breakable, skins}
\newcommand{\diff}[1]{{\color{blue!70!black}\rlap{\,$\downarrow$\scriptsize#1}}}
\definecolor{promptbg}{RGB}{248, 249, 252}
\definecolor{promptframe}{RGB}{180, 185, 200}
\definecolor{titlebg}{RGB}{70, 90, 120}
\definecolor{keywordcolor}{RGB}{248, 249, 252}
\definecolor{stringcolor}{RGB}{248, 249, 252}
\definecolor{commentcolor}{RGB}{130, 130, 140}

\newtcblisting{promptbox}[2][]{
    enhanced,
    breakable,
    colback=promptbg,
    colframe=promptframe,
    colbacktitle=titlebg,
    coltitle=white,
    fonttitle=\bfseries\small,
    title={#2},
    boxrule=0.6pt,
    arc=3pt,
    left=8pt,
    right=8pt,
    top=4pt,
    bottom=4pt,
    toptitle=3pt,
    bottomtitle=3pt,
    listing only,
    segmentation style={draw=none, line width=0pt},
    listing options={
        basicstyle=\ttfamily\scriptsize\linespread{1.1}\selectfont,
        backgroundcolor=\color{promptbg},
        breaklines=true,
        breakatwhitespace=false,
        columns=fullflexible,
        keepspaces=true,
        showstringspaces=false,
        commentstyle=\color{commentcolor}\itshape,
        keywordstyle=\color{keywordcolor},
        stringstyle=\color{stringcolor},
        inputencoding=utf8,
        extendedchars=true,
        xleftmargin=0pt,
        xrightmargin=0pt,
        frame=none, 
        framesep=0pt, 
        rulecolor=\color{promptbg},
        inputencoding=utf8,
        extendedchars=true,
        breakindent=0pt,
    },
    #1
}

\newtcolorbox{casestudybox}[2][]{
    enhanced,
    width=\textwidth,
    colback=promptbg,
    colframe=promptframe,
    colbacktitle=titlebg,
    coltitle=white,
    fonttitle=\bfseries\small,
    title={#2},
    boxrule=0.6pt,
    arc=3pt,
    left=8pt,
    right=8pt,
    top=6pt,
    bottom=6pt,
    toptitle=3pt,
    bottomtitle=3pt,
    #1
}

\title{MedInsightBench: Evaluating Medical Analytics Agents Through Multi-Step Insight Discovery in Multi-modal Medical Data}

\author{
\textbf{Zhenghao Zhu\textsuperscript{1,*}},
\textbf{Chuxue Cao\textsuperscript{1,*}},
\textbf{Yuanfeng Song\textsuperscript{2,$\dagger$}},
\textbf{Xing Chen\textsuperscript{2}},
\\
\textbf{Caleb Chen Cao\textsuperscript{1}},
\textbf{Sirui Han\textsuperscript{1,$\dagger$}},
\textbf{Yike Guo\textsuperscript{1}}
\\
\textsuperscript{1}The Hong Kong University of Science and Technology, Hong Kong, China\\
\textsuperscript{2}ByteDance, China \\
}

\begin{document}
\maketitle

\renewcommand{\thefootnote}{\fnsymbol{footnote}}
\footnotetext[1]{These authors contributed equally to this work.} 
\footnotetext[2]{Corresponding authors.} 
\renewcommand{\thefootnote}{\arabic{footnote}}

\begin{abstract}

In medical data analysis, extracting deep insights from complex multi-modal pathological data is essential for enhancing patient care, diagnostic accuracy, and healthcare operations. Despite this clinical significance, current benchmarks fail to adequately evaluate Large Multi-modal Models (LMMs) on medical insight discovery. Therefore, we introduce MedInsightBench, the first benchmark comprising 332 curated cases from The Cancer Genome Atlas, each annotated with expert-derived insights. Instead of simple QA, this benchmark evaluates whether LMMs can derive clinically grounded insights tied to tumor staging, prognosis, and biomarker expression. Our initial analysis reveals that existing LMMs struggle with MedInsightBench, primarily due to a lack of domain expertise and an inability to perform multi-step reasoning. Consequently, we propose MedInsightAgent, an automated agent framework featuring three specialized modules: Visual Root Finder, Analytical Insight Agent, and Follow-up Question Composer. Extensive experiments confirm that MedInsightAgent effectively addresses the limitations of existing models, consistently demonstrating superior performance in discovering complex medical insights.

\end{abstract}

\section{Introduction}

Recent advancements in medical data analysis using large multi-modal models (LMMs) have significantly improved clinical diagnosis~\citep{alSaadmultimodal, sun2025medical, xu2024comprehensive}. Medical insight detection that transforms multi-modal pathological data, such as pathological images and associated textual information, into actionable insights, such as determining tumor staging, assessing margin status, and identifying nodal metastasis, is crucial to improve diagnostic accuracy, guide treatment decisions, and enable new scientific discoveries~\citep{zhan2025retrieval, lu2024multimodal}.

Despite recent strides in LMMs for combined visual–language reasoning~\citep{alSaadmultimodal, sun2025medical, xu2024comprehensive}, their efficacy in real-world clinical insight discovery remains limited~\citep{fan-etal-2025-ai, schmidgall2024agentclinic}. Current benchmarks primarily probe surface-level facts rather than complex clinical cognition, failing to address the need for identifying occult pathological patterns, linking image features to molecular signatures, or synthesizing complementary visual and textual evidence for prognosis~\citep{pandit2025medhallu, shang2025medrepbench, wu2025medcasereasoning, tang2025medagentsbench}. 
To bridge this gap, we propose \textbf{\textit{MedInsightBench}}, which evaluates LMM capacity to derive actionable clinical insights through a curated dataset of complex images, realistic analytical tasks, and expert-validated question-insight pairs.

\begin{table*}[htbp!]
\centering
\renewcommand{\arraystretch}{1.2}
\resizebox{\textwidth}{!}{
\begin{tabular}{lccccccc}
\toprule
\rowcolor{gray!20} 
\textbf{Dataset} 
& \textbf{Input} 
& \textbf{Output} 
& \textbf{Task Type}
& \textbf{Topic \& Area} 
& \textbf{Multi-Step} 
& \textbf{Size} 
& \textbf{Construction Method} \\
\midrule

Spider 2.0 \cite{leispider}          
& Question       
& SQL Query         
& Text-to-SQL
& Enterprise-level  
& Limited
& 632   
& Machine \& Human-Labeled \\

VisEval \cite{chen2024viseval}
& Question+Chart
& Evaluation
& Visual Understanding
& Data Visualization
& No
& 500+
& Human-Labeled \\

MatPlotBench \cite{yang2024matplotagent}        
& Question+Table 
& Vis Image    
& Visual Generation
& Data Visualization 
& Partial
& 100 
& Machine \& Human-Labeled \\ 

InfiAgent-DABench \cite{hu2024infiagent}   
& Question+Table 
& Answer      
& Data Analysis
& Data Analysis     
& Partial
& 603  
& Machine-Labeled \\

\rowcolor{blue!6}
InsightBench \cite{sahuinsightbench}        
& \cellcolor{blue!12}Goal+Table 
& \cellcolor{blue!12}Insights    
& Insight Discovery
& Business Analysis 
& \cellcolor{blue!12}Extensive
& 100 
& Human-Labeled \\

MedRepBench \cite{shang2025medrepbench}
& Image+Report
& Interpretation
& Medical Report Analysis
& Clinical Analysis
& Limited
& 1K+
& Human-Annotated \\

\midrule

\rowcolor{green!10}
\textbf{MedInsightBench} 
& \cellcolor{green!22}\textbf{Goal+Image} 
& \cellcolor{green!22}\textbf{Insights} 
& \cellcolor{green!22}\textbf{Medical Insight Discovery}
& \cellcolor{green!22}\textbf{Medical Analysis} 
& \cellcolor{green!22}\textbf{Extensive}
& \textbf{332} 
& \cellcolor{green!22}\textbf{Machine-Labeled \& Human-Verified} \\

\bottomrule
\end{tabular}
}
\caption{
Comparison of MedInsightBench with existing benchmarks.
Our benchmark is the first to evaluate goal-driven multi-step medical insight discovery from image-modal medical data.
}
\label{tab:dataset_table}
\end{table*}

Based on publicly available cancer pathology resources, MedInsightBench comprises 332 cases and 3,933 insights covering six categories. The benchmark focuses on paired pathology images and associated textual evidence, with TCGA serving as the primary source due to the availability of large-scale, high-quality pathology images and associated clinical information. Our methodology involves downsampling Whole Slide Images (WSIs) into high-resolution patches, pairing them with human-verified evidence snippets from diagnostic reports, and distilling logical analysis goals to guide structured insight generation. As summarized in Table~\ref{tab:dataset_table}, MedInsightBench fills an important gap in existing benchmarks by pairing pathology images with explicit analytical goals and expert-verified question--insight--evidence triplets, thereby enabling goal-driven, multi-step medical insight discovery rather than standard QA, SQL generation, or general visual reasoning. 

Despite our comprehensive evaluation of LMMs, the observed results on this benchmark expose systematic performance gaps in their insight discovery capabilities within pathology. These models struggle with complex clinical workflows that demand high-precision image parsing, domain-constrained statistical reasoning, and verifiable inference. The observed fragility in chain-of-thought stability and latent interpretability underscores the inherent limitations of LMM architectures when deployed in high-stakes clinical settings. To bridge these gaps, we propose \textbf{\textit{MedInsightAgent}}, a multi-agent framework designed to orchestrate three synergistic components: 1) \textbf{Visual Root Finder}, which anchors high-level diagnostic queries in specific image cues and domain knowledge; 2) \textbf{Analytical Insight Agent}, which performs region-specific scrutiny to generate grounded, evidence-based responses; and 3) \textbf{Follow-Up Question Composer}, which iteratively synthesizes intermediate findings into deeper, derivative insights. By facilitating structured information exchange and iterative verification, MedInsightAgent enables an exploratory, reliable, and interpretable paradigm for clinical reasoning.

In our experiments, we benchmark a diverse suite of LMMs and agent methods on MedInsightBench using a comprehensive and rigorous evaluation protocol, encompassing metrics such as Insight Recall, Precision, $F1$, and Novelty. The results highlight several key challenges in automated medical insight discovery and further demonstrate that our proposed MedInsightAgent significantly and consistently enhances the insight discovery performance of base LMMs across different settings. In summary, our main contributions are as follows:

\begin{itemize}[leftmargin=*,noitemsep,nolistsep]
\item We introduce MedInsightBench, a novel multi-modal pathological benchmark advancing automated medical insight discovery. By pairing high-resolution pathology images with clinical narratives, it leverages hierarchical reasoning tasks and validated metrics to rigorously assess the depth and clinical utility of generated knowledge.
\item We propose a multi-agent collaborative framework engineered for automated, high-fidelity medical insight discovery. The framework formalizes agent roles and interaction protocols to effectively synthesize local visual analysis, context-aware cross-sample inference, and externally retrieved domain knowledge.
\item We provide extensive empirical analysis of several baseline LMMs and agentic methods. The experiments prove the benchmark's discriminative power and demonstrate that our multi-agent approach improves the precision and interpretability of extracted clinical information.
\end{itemize}

\begin{figure*}[t!]
    \centering
    \includegraphics[width=\linewidth]{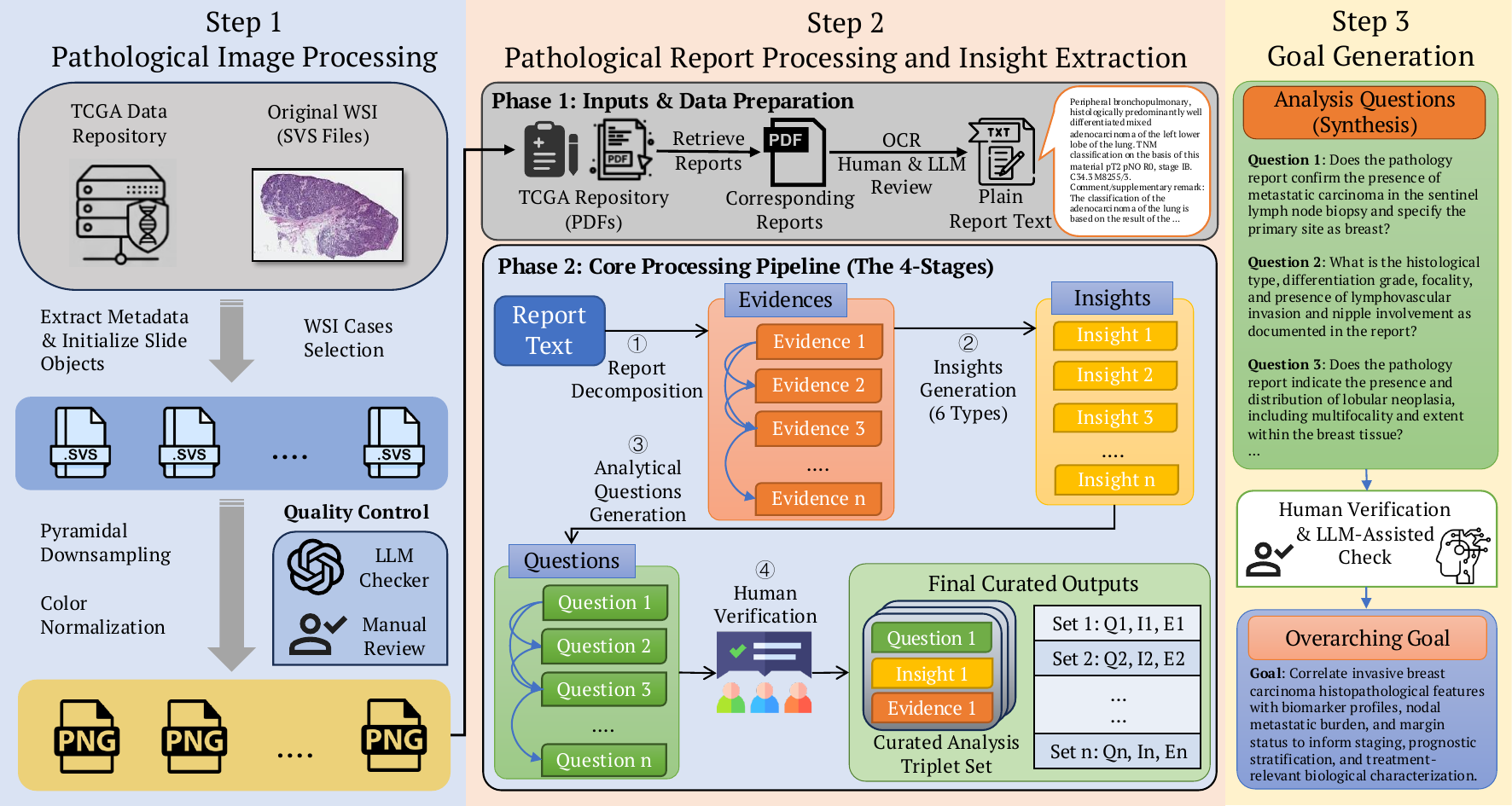}
    \caption{The dataset construction pipeline of MedInsightBench. This pipeline consists of 3 steps: 1) \textbf{Pathological Image Processing}. WSIs are standardized and quality-checked. 2) \textbf{Report Processing \& Insight Extraction}. Reports are converted to text, insights and questions are generated, and verified by experts. 3) \textbf{Goal Generation}. An overarching analysis goal is synthesized from the questions and validated for guiding the analysis.}
    \label{fig:data_construction_pipeline}
    \vspace{-7pt}
\end{figure*}

\section{Medical Insight Benchmark}

\subsection{Preliminary Study of Insight Discovery}

In the insight discovery task, there are a variety of viable approaches. For tabular data insights, multi-agent pipelines have proven particularly effective. Two state-of-the-art multi-agent paradigms are Pérez et al.~\citep{perez2025llm}, which uses SQL to extract information from structured tables, and Agent Poirot~\citep{sahuinsightbench}, which relies on Python scripts and standard data-analysis libraries to retrieve the relevant statistics and evidence. 

To facilitate clinical insight discovery in image-modal data, we evaluate both LMMs and agentic framework paradigms. Given an image and a predefined diagnostic goal, LMM generates analytical insights through direct, end-to-end multimodal reasoning. In contrast, the multi-agent pipeline adopts a staged methodology: it systematically decomposes the task into goal-directed questions, resolves them through iterative reasoning or targeted tool invocation, and synthesizes these intermediate findings into a coherent, clinically actionable insight.

Guided by InsightBench~\citep{sahuinsightbench}, we identified and implemented three key design requirements to construct our novel medical image-based insight discovery benchmark:

\begin{enumerate}[leftmargin=*,noitemsep,nolistsep]
\item \textbf{Medical image quality and completeness}: 
The images must clearly and fully depict the target content, ensuring that all relevant features are sufficiently visible and distinguishable for accurate interpretation and downstream analysis.
\item \textbf{Explicit analytical goal}: 
The analysis goal must be strictly unambiguous and clearly articulate the intended focus, delineating the specific comparison metrics, evaluation axes, or key dimensions of analysis to be considered.
\item \textbf{Question–Insight consistency}: 
Each insight must be supported by a clear, well-formed question and grounded in solid, verifiable evidence. The questions should be comprehensive and multidimensional, while the insights should be meaningful, informative, and well-rounded.
\end{enumerate}

\subsection{Data Construction}

From our preliminary study, we identified the key priorities and objectives for dataset construction: (i) high-quality and comprehensive medical images; (ii) an explicit and well-specified analytical goal; (iii) comprehensive, in-depth, multi-dimensional exploratory insights. Among publicly available medical datasets, the Cancer Genome Atlas (TCGA) provides various types of cancer and associated patient samples. Each sample includes tumor-associated images and paired pathology reports, which align well with our requirements. Therefore, we select it as our source data.

Our construction methodology combines mainly manual curation with LLM–assisted generation, supplemented by a human review to verify image quality and evidence validity (Appendix~\ref{appendix:human_annotation_details}). The overall pipeline is illustrated in Figure \ref{fig:data_construction_pipeline}. We describe each step of the pipeline in detail as follows.

\paragraph{Step 1: Pathological Image Processing} In the TCGA repository, each pathological whole-slide image (WSI) is stored as SVS. To convert WSI into suitable inputs for LMMs, we applied a standardized image processing pipeline. First, we instantiate a slide object and extract essential metadata such as pixel spacing and the dimensions of each pyramid level. Next, to preserve the global structure and large-scale morphological features while reducing the WSI to an acceptable size, we perform whole-slide downsampling. Given a target maximum output dimension, we compute an appropriate downsample ratio and select the optimal pyramid level. After color normalization, the images are exported as PNGs for downstream use.

To ensure that the final images are clear, complete and usable, we also utilize an automated check using an LMM combined with manual review. This step filters out images that are unreadable or corrupted and yields a curated set of pathological images suitable for data analysis.

\paragraph{Step 2: Pathological Report Processing and Insight Extraction}
We first retrieve the corresponding pathology reports (PDF) from the TCGA repository based on the case name of each sample. Next, we convert the reports to plain text through OCR and then inspect and correct them through carefully designed LLM assistance and rigorous human verification. We inspect each report based on multiple quality criteria, including the absence of invalid characters or corrupted content, coherence of diagnostic statements, and overall alignment between textual descriptions and expected clinical content. Therefore, insight generation from plain-text reports is carried out in four stages as follows:

\begin{enumerate}[leftmargin=*,noitemsep,nolistsep]

\item \textbf{Report Decomposition}: 
We apply an LLM to extract the key items of the report, represented as a sequence of evidence snippets that form a progressive, interrelated chain of findings.
\item \textbf{Insight Generation}: Based on 6 types of insights (Details in Appendix~\ref{sec:insight_types}), we analyze each evidence and generate insights. Moreover, we compute a confidence score for each insight to indicate quality, and insights with low confidence are manually filtered.
\item \textbf{Analytical Questions Generation}: 
To enhance analytical depth and hierarchy, we deliberately pose goal-directed questions for each insight, ensuring a clear and logical progression that systematically guides the analysis and enables incremental discovery of deeper, more meaningful, and well-grounded findings.
\item \textbf{Human Verification}: 
To ensure clinical consistency and factual alignment, we conducted a structured human evaluation of the triplets (comprising question, insight, and evidence), thereby preventing potential over-extrapolation.
\end{enumerate}

\paragraph{Step 3: Goal Generation} The analytical goal has two key properties: (i) it must be clearly and unambiguously stated. (ii) effectively guide both the generation of analysis questions and the overall analytical strategy. 
We analyze the logical relationships and dependencies among these generated questions and synthesize a concise, overarching analysis goal. 
To prevent hallucinations, misinterpretations, or post-hoc synthesis artifacts, each generated goal is strictly vetted through our human-in-the-loop validation constraints. We selected goals that are precise, coherent with the underlying questions, and conducive to downstream analyses.

\subsection{Data Quality Analysis}
\label{sec:data_quality_ana}
To verify the quality of the dataset, we conducted an in-depth annotation across three dimensions:

\begin{itemize}[leftmargin=*,noitemsep]
\item \textbf{Correctness}: whether each set of questions strictly corresponds to the stated goal and the pathological images without factual errors.
\item \textbf{Rationality}: whether each insight satisfies the goal's requirements and is logically sound. 
\item \textbf{Coherence}: whether insights in each case are internally consistent and mutually compatible.
\end{itemize}

\begin{figure}[htbp!]
    \centering
    \includegraphics[width=\linewidth]{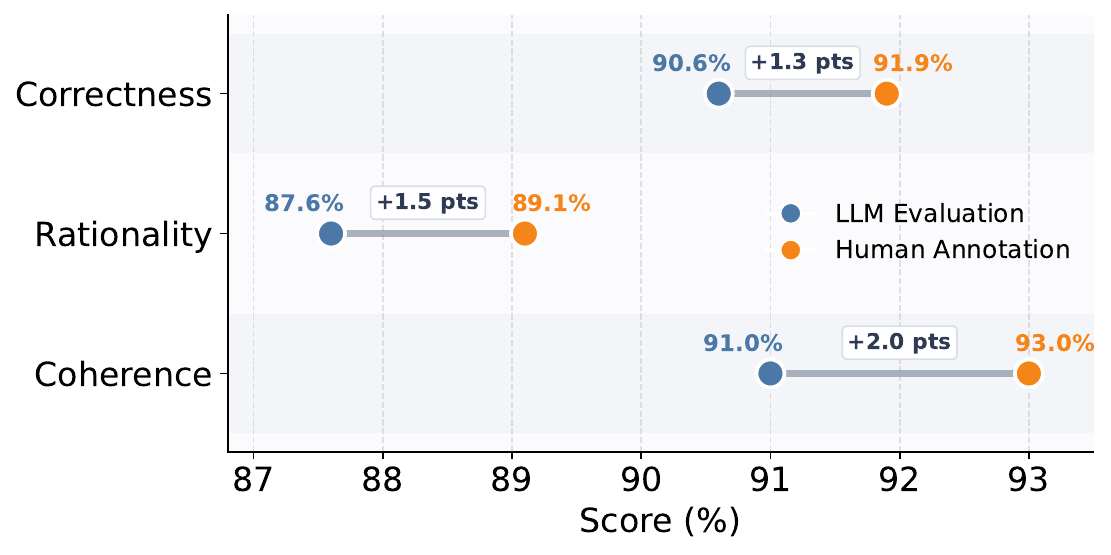}
    \caption{Quality Comparison Between LLM Evaluation and Human Annotation.}
    \label{fig:quality_comparison}
\end{figure}

\begin{figure}[htbp!]
    \centering
    \includegraphics[width=\linewidth]{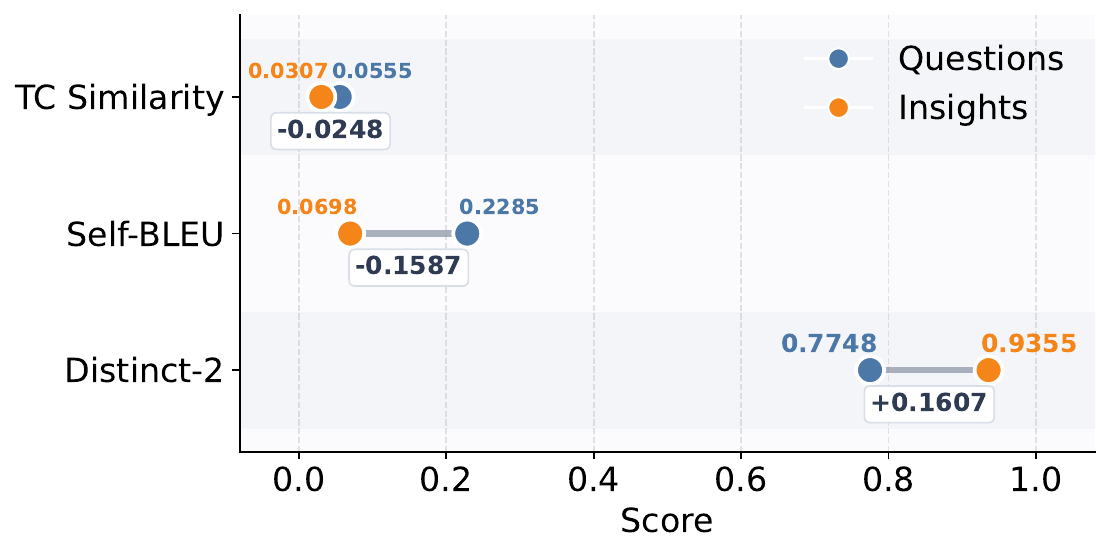}
    \caption{Comparison of average TF‑IDF cosine (TC) similarity, Self‑BLEU, and Distinct‑2 diversity scores between Question and Insight in MedInsightBench.}
    \label{fig:redund_assessment}
\end{figure}

We randomly sampled 100 representative cases and annotated them by both LMM (i.e., OpenAI o3~\citep{openai2025competitiveprogramminglargereasoning}) and human experts, computing the accuracy rate for each dimension. The results are reported in Figure~\ref{fig:quality_comparison}. In addition, we evaluated redundancy for each question and insight using three complementary metrics. First, we compute cosine similarity based on TF-IDF vector representations and average the resulting scores. Second, we computed Self-BLEU~\citep{zhu2018texygen} for each sentence to assess n-gram repetitiveness. Third, we measured Distinct-2~\citep{li-etal-2016-diversity}, defined as the ratio of unique bigrams to total bigrams across all sentences. Generally, a higher TF-IDF cosine similarity and Self-BLEU indicate greater redundancy, while a Distinct-2 value closer to 1 reflects greater lexical diversity and lower redundancy. Figure~\ref{fig:redund_assessment} reports these redundancy statistics. Through this quality assurance process, our dataset meets a high standard of reliability and validity.

\subsection{Evaluation Framework}

We evaluate medical insight discovery using four complementary metrics: Insight Recall, Insight Precision, Insight $F1$, and Insight Novelty. Recall measures whether the model covers the key evidence-backed insights that should be discovered from the pathology image and report; Precision measures whether the generated insights are factually grounded and free from over-extrapolation; $F1$ balances completeness and faithfulness; and Novelty captures whether the model can produce additional clinically meaningful insights beyond the annotated ground truth while remaining consistent with the available evidence. Because medical insight discovery is inherently open-ended, no single metric is sufficient on its own: Recall alone may favor conservative outputs, whereas Novelty alone may reward plausible but weakly supported claims. Therefore, we treat Insight $F1$ as the primary metric for faithful pathology-grounded discovery, and use Novelty as a complementary indicator of exploratory capability. The formal definitions of all metrics are provided in Appendix~\ref{appendix:evaluation_details}.

\begin{figure*}[tbp!]
    \centering
    \includegraphics[width=\linewidth]{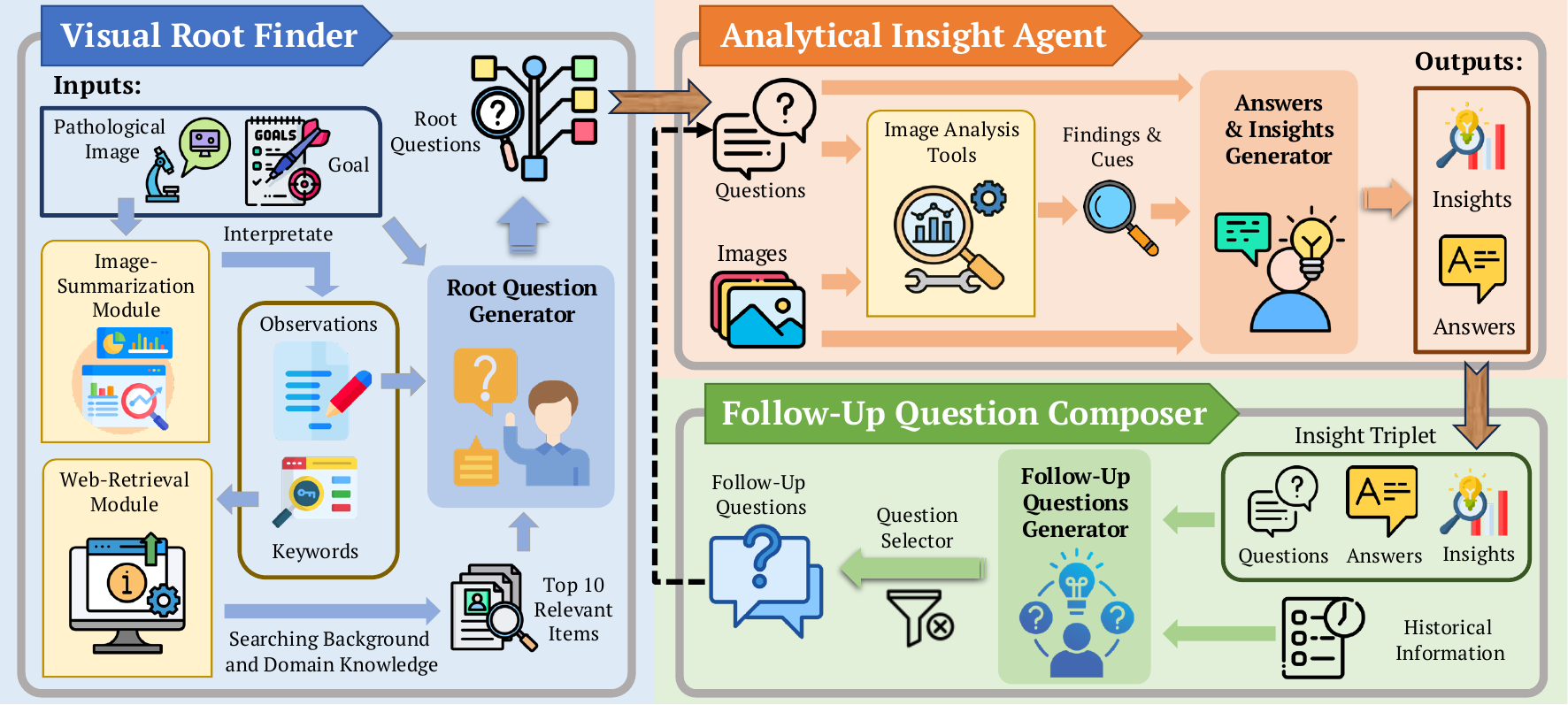}
    \caption{Overview of our proposed agentic framework, \textit{MedInsightAgent}, for goal-driven medical insight discovery. It consists of three collaborative components: (1) \textbf{Visual Root Finder} for image summarization and root-question generation; (2) \textbf{Analytical Insight Agent} for grounded question-guided analysis; and (3) \textbf{Follow-Up Question Composer} generates deeper follow-up questions for progressive multi-step medical reasoning.}
    \label{fig:MedAgentWorkflow}
\end{figure*}

\subsection{Benchmark Statistic}

The MedInsightBench dataset comprises 332 samples, each containing a single cancer pathology image, a specific goal, and several medical insights, yielding a total of 3,933 insights across the dataset. Each sample is annotated with one of the four difficulty levels. Furthermore, each insight is labeled with an insight category, an associated question, and an excerpt of evidence drawn from the report. Detailed statistics are deferred to Appendix~\ref{appendix:dataset}.

\section{The Proposed MedInsightAgent}

Due to the suboptimal performance and inherent limitations of LMM in medical insight discovery tasks, we design a novel multi-agent framework named \textbf{\textit{MedInsightAgent}}. The framework decomposes the insight discovery process into three specialized agents: 1) \textbf{Visual Root Finder}: Given the analytical goal, analyze the image, summarize, identify salient visual features, and generate an initial set of root questions. 2) \textbf{Analytical Insight Agent}: Answer each question using the image and associated evidence, and finally generate meaningful medical insights. 3) \textbf{Follow-up Question Composer}: Generate follow-up questions that probe deeper or explore complementary perspectives to refine and extend the discovered insights. The overall architecture is illustrated in Figure~\ref{fig:MedAgentWorkflow}, with the comprehensive pseudocode presented in Appendix~\ref{appendix:algorithm}. We describe the processing flow and implementation details of each agent as follows.

\subsection{Visual Root Finder}

The Visual Root Finder ($\mathrm{VRF}$) takes a medical image $I$ and an analytical goal $G$ as input and generates an initial set of root questions $Q = \{q_i\}_{i=1}^m$, where $m$ is the number of questions. These root questions define the primary directions for exploration and guide subsequent insight generation.

To improve the quality of root questions, Visual Root Finder first gathers supplementary information. It incorporates two information-acquisition modules: (1) \textbf{Image-Summarization Module} $\mathcal{ISM}_{\text{img}}$. To broadly explore the medical image, the module performs an initial interpretation, extracting prominent visual features and observations $F$. Then it generates various and representative keywords $K$, formalized as $\mathcal{ISM}_\text{img}: I \mapsto (F, K)$. (2) \textbf{Web-Retrieval Module} $\mathcal{WRM}$. Using keywords $K$, this module retrieves domain knowledge by querying online resources (e.g., literature, reports) and returns the top ten relevant items $D \;=\; \mathcal{WRM}(K) \;=\; \{d_1,\dots,d_{10}\}$.

Finally, the \textbf{Root Question Generator} $\mathcal{L}$ takes the medical image, the predefined analysis goal, and the retrieved information to produce a set of high-quality and concrete root questions that form the foundation for downstream analytical agents. The general process is formalized in Eq.~\ref{VRF_equation}.

\begin{equation}
\label{VRF_equation}
\mathrm{VRF}(I,G) \;=\; \mathcal{L} \big( I,G,F,D \big) \;\Rightarrow\; Q
\end{equation}

\subsection{Analytical Insight Agent}

The Analytical Insight Agent ($\mathrm{AIA}$) generates answers $A = \{a_i\}_{i=1}^m$ of the root questions and derives meaningful insights $S = \{s_i\}_{i=1}^m$. Since different questions probe distinct analytical facets, directly interpreting the pathological image often leads to hallucinations or incomplete responses. Thus, it is essential to explicitly extract image evidence relevant to questions before answering them.

For this targeted evidence extraction, we employ PathGen-LLaVA \citep{sun2025pathgenm}, which is an LMM built on the LLaVA architecture and further fine-tuned on the PathGen pathology dataset \citep{sun2025pathgenm}, as an \textbf{Image-Analysis Tool} $\mathcal{IAT}$. For each root question $q_i$, it carefully analyzes the image $I$ and outputs relevant pathological findings along with detailed visual cues $E_i$, where $E_i = \mathcal{IAT}(I, q_i)$. These structured and question-specific image features serve as grounded and reliable evidence for subsequent reasoning.

Finally, the \textbf{Answers and Insights Generator} $\mathcal{G}$ takes the question, the pathological image, and the extracted findings to produce a rational answer and a concise, clinically meaningful insight. The overall formula is shown in Eq.~\ref{AIA_equation}.

\begin{equation}
    \label{AIA_equation}
    \mathrm{AIA}(I,Q)\;=\; \Big\{\, \mathcal{G}\big(q_i,I,E_i \big) \Big\}_{i=1}^m \;\Rightarrow\; A,S
\end{equation}

\subsection{Follow-up Question Composer}
The initial set of root questions often exhibits inherent limitations, including insufficient coverage of pathological features and significant stochastic variability. To address this, we introduce the Follow-up Question Composer ($\mathrm{FQC}$), which designed to derive more comprehensive and nuanced inquiries from each primary root question. The generated set of follow-up questions $T = {t_i}_{i=1}^m$ is strictly governed by two foundational criteria: (i) They must be relevant to the image $I$ and fully aligned with the analytical goal $G$. (ii) They must be distinct yet logically derived from the original root question, effectively extending the inquiry to explore additional facets of the pathological image.

The \textbf{Follow-Up Question Generator} $\mathcal{F}$ initially generates a set of $n$ candidate follow-up questions, denoted as $C = {c_i}{i=1}^n$. Subsequently, the \textbf{Question Selector} $\mathcal{S}$ evaluates and scores each candidate question in this set and then selects the one with the highest score, referred to as $c_{best}$. This overall process is formally described in Eq.~\ref{FQC_equation}.

\begin{equation}
\label{FQC_equation}
\begin{aligned}
    \mathrm{FQC}(I,G,Q,A,F,D) = \mathcal{S}(C) \;\Rightarrow\; c_{\text{best}} \\
    \text{where} \quad C = \big\{\mathcal{F}(I,G,q_i,a_i,F,D) \big\}_{i=1}^n
\end{aligned}
\end{equation}

The selected follow-up question $c_{best}$ is then passed to the Analytical Insight Agent to generate new insights. The process is controlled by an exploration depth parameter $p\;(p \geq 0)$, which specifies the number of follow-up iteration cycles. Each root question is expanded through $p$ rounds before termination, after which the system outputs all accumulated insights. In particular, if $r$ root questions are generated, the total number of final insights ($Ins$) can be computed as $Ins = r \times (q+1)$.

\section{Experiments and Analysis}
\subsection{Experimental Setup}
\begin{table*}[htbp!]
\centering
\fontsize{8.0pt}{8.0pt}\selectfont 
\setlength{\tabcolsep}{3.8pt}
\renewcommand{\arraystretch}{1.12}
\resizebox{\textwidth}{!}{
\begin{tabular}{l|cc|cc|cc|cc}
\toprule
\multirow{2}{*}{\textbf{Baselines}} & \multicolumn{2}{c|}{\textbf{Insights Recall}} & \multicolumn{2}{c|}{\textbf{Insights Precision}} & \multicolumn{2}{c|}{\textbf{Insights $F_1$}} & \multicolumn{2}{c}{\textbf{Insights Novelty}}\\
& ROUGE-1~~&~G-Eval & ROUGE-1~~&~G-Eval & ROUGE-1~~&~G-Eval & Original~~&~Innovation\\
\midrule
\rowcolor{lightgray!50}
\multicolumn{9}{l}{\textbf{\textit{Direct Decoding} (LMMs-Only)}} \\
GPT-4o & 0.1799 & 0.2978 & 0.2091 & 0.3577 & 0.1934 & 0.3250 & 0.1293 & 0.2094 \\
GPT-5 & 0.1866 & 0.3052 & 0.1851 & 0.3649 & 0.1858 & 0.3324 & 0.1322 & 0.2131 \\
Deepseek-VL2 & 0.1816 & 0.3226 & \textbf{0.2274} & 0.4070 & 0.2019 & 0.3599 & 0.1959 & 0.2712 \\
InternVL3-38B & 0.1774 & 0.3390 & 0.2002 & 0.3993 & 0.1881 & 0.3667 & 0.1607 & 0.2553 \\
Qwen2.5-VL-32B-Instruct & 0.1917 & 0.3975 & 0.2139 & 0.4848 & 0.2022 & 0.4368 & 0.3493 & 0.4171 \\
Qwen3-VL-32B-Instruct & \textbf{0.2064} & \textbf{0.4295} & 0.2267 & \textbf{0.5372} & \textbf{0.2170} & \textbf{0.4774} & \textbf{0.3531} & \textbf{0.4488} \\
PathGen-LLaVA & 0.1849 & 0.3420 & 0.1989 & 0.3845 & 0.1916 & 0.3620 & 0.1893 & 0.2936 \\
\midrule
\rowcolor{lightgray!50}
\multicolumn{9}{l}{\textbf{\textit{Agent Methods} (Qwen3-VL-32B-Instruct as backbone)}} \\
Chain-of-Thought (CoT) & 0.2053 & 0.4379 & 0.2281 & 0.5374 & 0.2161 & 0.4826 & 0.3549 & 0.4583 \\
ReAct & 0.2092 & 0.4377 & 0.2344 & 0.5390 & 0.2211 & 0.4831 & 0.3492 & 0.4456 \\
ReAct + same tools & 0.2105 & 0.4562 & 0.2397 & 0.5698 & 0.2242 & 0.5067 & 0.3853 & 0.4679 \\
Plan-and-Solve & 0.2117 & 0.4534 & 0.2378 & 0.5633 & 0.2240 & 0.5024 & 0.3749 & 0.4613 \\
Self-Refine & 0.2080 & 0.4467 & 0.2342 & 0.5554 & 0.2203 & 0.4952 & 0.3584 & 0.4493 \\
Reflexion & 0.2125 & 0.4573 & 0.2395 & 0.5670 & 0.2252 & 0.5063 & 0.3826 & 0.4548 \\
\rowcolor{blue!20} MedInsightAgent & \textbf{0.2233} & \textbf{0.4703} & \textbf{0.2483} & \textbf{0.5921} & \textbf{0.2351} & \textbf{0.5242} & \textbf{0.4274} & \textbf{0.4995} \\
\bottomrule
\end{tabular}
}
\caption{Performance comparison of baseline models on the MedInsightBench across insight extraction metrics.}
\label{tab:main_results}
\end{table*}

\paragraph{Baselines.} We conduct extensive experiments on various competitive baselines:

\begin{itemize}[leftmargin=*,noitemsep,nolistsep]
\item \textbf{Large Multi-modal Models.} 
We directly utilize several LMMs including GPT-4o~\citep{openai2024gpt4ocard}, GPT-5~\citep{openai2025gpt5card}, Deepseek-VL2~\citep{wu2024deepseek}, InternVL3-38B~\citep{chen2024internvl}, Qwen2.5-VL-32B-Instruct~\citep{bai2025qwen25vltechnicalreport}, Qwen3-VL-32B-Instruct~\citep{qwen3technicalreport} and PathGen-LLaVA~\citep{sun2025pathgenm} in Image-Analysis Tool for insights.

\item \textbf{Agentic Methods.}
We compare against representative agentic baselines, including CoT~\citep{wei2022chain}, ReAct~\citep{yao2023react}, Plan-and-Solve~\citep{wang2023plan}, Self-Refine~\citep{madaan2023selfrefine}, and Reflexion~\citep{shinn2023reflexion}, to examine whether explicit reasoning, planning, tool use, and self-refinement can improve medical insight discovery beyond direct decoding. Detailed method descriptions and implementations are provided in Appendix~\ref{appendix:details_imple}.

\end{itemize}

\paragraph{Implementation.}
To maintain consistency, we utilized the Qwen3-VL-32B-Instruct~\citep{qwen3technicalreport} across all agentic baselines. Additionally, MedInsightAgent conducted four iterations with three new questions per round. The same total question budget was applied to all other agentic baselines to ensure a fair comparison.

\paragraph{Metrics.} For recall and precision assessment, we employ two evaluators: ROUGE‑1~\citep{lin2004rouge} and G-Eval~\citep{liu2023g}. Specifically, the G-Eval score is calculated as the average of the GPT‑3.5 Turbo~\citep{openai2023gpt35turbo} and Gemini 2.5 Pro~\citep{comanici2025gemini25pushingfrontier} scores. Appendix~\ref{appendix:evaluation_details} provides the full details of all metric calculations.

\subsection{Experimental Results and Findings}

\paragraph{MedInsightAgent achieves the best overall performance.}
Table~\ref{tab:main_results} reports the results on MedInsightBench. Among the direct-decoding LMMs, Qwen3-VL-32B-Instruct is the strongest baseline overall, achieving the best ROUGE-1 recall and F$_1$, as well as the highest G-Eval recall, precision, and F$_1$. DeepSeek-VL2 obtains the highest ROUGE-1 precision among LMM-only baselines, but remains behind Qwen3-VL-32B-Instruct on most other metrics. Building on this backbone, MedInsightAgent further improves performance across all evaluation dimensions, yielding the best extraction quality and novelty scores. These results show that our framework not only retrieves more relevant insights, but also produces more informative discoveries.

\paragraph{Agentic orchestration contributes beyond tool access.}
Compared with prompt-based agent baselines built on the same Qwen3-VL-32B-Instruct backbone, MedInsightAgent consistently outperforms Chain-of-Thought, ReAct, Plan-and-Solve, Self-Refine, and Reflexion across all evaluation metrics. The gap remains even when ReAct is equipped with the same external tools used in our system. Although tool access improves ReAct over the vanilla version, it still falls short of MedInsightAgent, suggesting that the gain comes not only from tools but also from how reasoning, tool use, and refinement are coordinated. This indicates that effective medical insight discovery depends on structured agent orchestration rather than one-step prompting with external capabilities.

\begin{table*}[tbp!]
\centering
\setlength{\tabcolsep}{3.8pt}
\renewcommand{\arraystretch}{1.0}
\resizebox{\textwidth}{!}{
\begin{tabular}{l|cccc}
\toprule
\textbf{Methods} & \textbf{Insights Recall} & \textbf{Insights Precision} & \textbf{Insights $F_1$} & \textbf{Insights Novelty} \\
\midrule
Direct Decoding (Qwen3-VL-32B-Instruct) & 0.4295 & 0.5372 & 0.4774 & 0.4488 \\
\midrule
MedInsightAgent (Qwen3-VL-32B-Instruct) & \textbf{0.4703} & \textbf{0.5921} & \textbf{0.5242} & \textbf{0.4995} \\
\ \ \ \ \ \ \ \ w/o Image-Summarization Module & 0.4615\diff{0.0088} & 0.5764\diff{0.0157} & 0.5126\diff{0.0116} & 0.4780\diff{0.0215} \\
\ \ \ \ \ \ \ \ w/o Web-Retrieval Module       & 0.4623\diff{0.0080} & 0.5739\diff{0.0182} & 0.5121\diff{0.0121} & 0.4705\diff{0.0290} \\
\ \ \ \ \ \ \ \ w/o Image-Analysis Tool        & 0.4537\diff{0.0166} & 0.5604\diff{0.0317} & 0.5014\diff{0.0228} & 0.4842\diff{0.0153} \\
\cmidrule(lr){1-1} \cmidrule(lr){2-5}
\ \ \ \ w/o Follow-Up Question Composer  & 0.4391\diff{0.0312} & 0.5511\diff{0.0410} & 0.4888\diff{0.0354} & 0.4653\diff{0.0342} \\
\bottomrule
\end{tabular}
}
\caption{Effect of each method and module within the MedInsightAgent framework. The blue arrows and values indicate the performance drop compared to the full MedInsightAgent(Qwen3-VL-32B-Instruct).}
\label{tab:ablation_study_results}
\end{table*}

\paragraph{Precision is consistently higher than recall.}
Insight precision consistently exceeds recall across all baselines under ROUGE-1 and G-Eval, suggesting that models are conservative. While this tendency to generate high-confidence, well-supported insights minimizes weakly grounded claims, it also limits exploratory depth and evidence coverage in pathology reports. The result highlights the inherent trade-off between correctness and comprehensiveness in medical insight discovery.

\paragraph{Semantic quality is better captured than lexical overlap.}
G-Eval scores substantially exceed ROUGE-1 across all methods. This gap suggests that surface-level overlap underestimates semantic quality, as many insights are expressed via non-reference wording. It also highlights the value of a multi-faceted evaluation protocol: ROUGE reflects lexical faithfulness, while G-Eval captures whether insights are clinically meaningful and well-formed.

\paragraph{Stronger extraction quality is generally associated with better novelty.}
Methods that achieve higher extraction scores also tend to obtain higher novelty scores, indicating that novelty in this setting is driven by better evidence selection and synthesis rather than unconstrained generation. In particular, the strongest models improve both the Original and Innovation novelty scores, suggesting that stronger reasoning can surface additional informative content beyond the initial extraction. This pattern supports the view that novelty and accuracy are complementary rather than conflicting goals.

\paragraph{External tools help, but orchestration matters more.}
ReAct, with the same tools as MedInsightAgent, shows clear gains over vanilla ReAct, confirming that external tools are useful for medical insight discovery. However, the remaining gap to MedInsightAgent suggests that tool access alone is insufficient. Our results indicate that the main benefit comes from structured task decomposition, iterative coordination, and role-specific refinement, which make the agent process more effective than a generic tool-augmented prompting strategy.

\begin{figure}
    \centering
    \includegraphics[width=\linewidth]{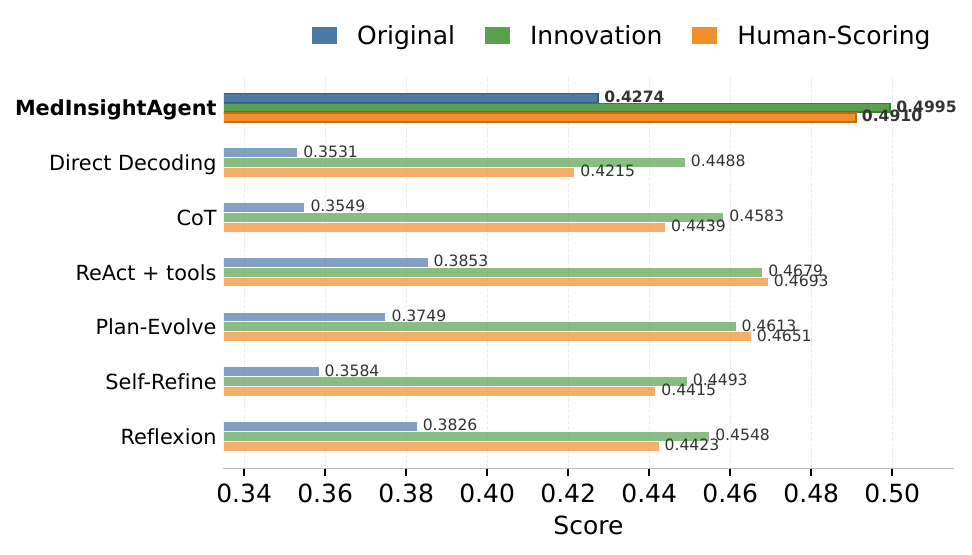}
    \caption{Comparison of novelty scores between model-generated and human-assessed insights across baselines.}
    \label{fig:novelty_comparison}
\end{figure}

\paragraph{The Novelty metric captures an agent's ability to surface previously unseen medical insights.}
As shown in Figure~\ref{fig:novelty_comparison}, MedInsightAgent consistently outperforms all baselines across automated novelty evaluation, innovation assessment, and human scoring. Notably, the improvements remain evident under human evaluation, indicating that the discovered insights are not only novel but also clinically meaningful. Although automated novelty scores are generally more optimistic in absolute magnitude, they remain well aligned with human judgments and preserve the relative ranking of methods.

\subsection{Ablation Study}

To evaluate the contribution of each module within MedInsightAgent, we conducted an ablation study in Table~\ref{tab:ablation_study_results}. Performance consistently declined upon the removal of any individual component, confirming the necessity of each module. Notably, the Image-Analysis Tool yielded the most significant impact on intrinsic quality metrics (Recall, Precision, and $F1$), as it facilitates targeted, goal-directed analysis of medical slides to provide critical evidence. Conversely, the absence of the Web-Retrieval Module triggered a sharp drop in insight-novelty scores, underscoring its pivotal role in integrating external domain knowledge to foster innovative discoveries. Furthermore, ablations on the Follow-up Question Composer reveal that multi-round iterative questioning is essential for deeper exploration and sustained improvement in insight novelty. Collectively, these results demonstrate that the synergy between visual analysis, external knowledge retrieval, and iterative interaction is fundamental to the agent's capability for generating comprehensive and high-quality medical insights.

\section{Conclusion}

We propose MedInsightBench, a novel and comprehensive benchmark for the rigorous, systematic, and precise evaluation of medical-insight discovery. This benchmark not only supports fully automated evaluation but also demonstrates strong and consistent concordance with human expert judgments. In addition, we introduce MedInsightAgent, a multi-agent framework that integrates multiple data-acquisition modules, analysis components, and external tools specifically designed for mining valuable insights from complex pathological images. Experimental results show that MedInsightBench reveals numerous key challenges and bottlenecks in medical-insight discovery, while MedInsightAgent effectively enhances the performance of several leading LMMs. 
In future work, we plan to further refine and optimize this multi-agent framework to substantially improve its capability in insight discovery. Meanwhile, building on the modality-agnostic benchmark construction framework, we will explore extending MedInsightBench beyond pathology to other medical domains, such as radiology and EHR-based analysis, when suitable paired data and structured insight annotations become available.

\section*{Limitations}
While MedInsightBench and MedInsightAgent establish a foundation for multi-step medical insight discovery, several limitations warrant future exploration: (1) Limited generalizability. Our reliance on TCGA pathology records restricts applicability to other modalities (e.g., radiology) and may neglect the demographic diversity essential for clinical practice. Future work will integrate multi-modal data and diverse patient cohorts to enhance robustness. (2) Methodological bias. While our construction leveraged LLM-assisted generation, this may introduce specific reasoning biases. Furthermore, metrics like Insight $F1$ and Novelty, while scalable, remain approximations; as shown in Appendix~\ref{appendix:evaluation_details}, token-level variations can lead to the overestimation of novelty. Consequently, these metrics should complement, not substitute, expert human judgment. (3) Lack of clinical validation. Our framework assesses benchmark performance rather than direct patient outcomes. Establishing a causal link between model-generated insights and improved clinical decision-making will require prospective, real-world trials. Finally, current downsampling of whole-slide images may overlook fine-grained microstructural features, necessitating higher-resolution analysis in future iterations.

\section*{Acknowledgments}
This work is funded in part by the HKUST Start-up Fund (R9911), Theme-based Research Scheme grant (T45-205/21-N), the InnoHK initiative of the Innovation and Technology Commission of the Hong Kong Special Administrative Region Government, and the research funding under HKUST-DXM AI for Finance Joint Laboratory (DXM25EG01).

\section*{Ethical Statement}
This work presents a medical dataset and a specialized agent designed to assist in clinical information analysis. We acknowledge the following ethical considerations:

\begin{enumerate}[leftmargin=*,noitemsep,nolistsep]
    \item  Decision Support, Not Diagnosis: The proposed agent system is intended as a decision-support tool to augment clinical workflows and should not be construed as a provider of definitive medical diagnosis or treatment advice.

    \item Human-in-the-loop: All outputs generated by the agent must be critically evaluated by qualified healthcare professionals. Final clinical decisions remain the sole responsibility of the human practitioner.

    \item Risks of Hallucinations: Despite rigorous training, the agent may occasionally generate incorrect or biased medical information (i.e., hallucinations). It should not be used in high-stakes clinical settings without expert verification.
\end{enumerate}

\bibliography{custom}

\clearpage
\appendix

\section{Related Works}

\subsection{Medical Data Analysis}

Recent research has introduced benchmarks and frameworks for evaluating large language models (LLMs) and agent systems on medical reasoning and data analysis tasks. Several datasets emphasize multi-step clinical reasoning and multi-modal expert questions, including MedAgentsBench~\citep{tang2025medagentsbench}, MedAgentBench~\citep{jiang2025medagentbench}, MedCaseReasoning~\citep{wu2025medcasereasoning}, MedXpertQA~\citep{zuo2025medxpertqa}, and the Chinese CMB~\citep{wang-etal-2024-cmb}. Comparative studies have also comprehensively evaluated emerging models, such as DeepSeek, across clinical reasoning and diverse medical tasks~\citep{tordjman2025comparative}. Other work addresses interactive clinical workflows and multi-agent collaboration. AI Hospital~\citep{fan-etal-2025-ai}, AgentClinic~\citep{schmidgall2024agentclinic}, 3MDBench~\citep{sviridov20253mdbench}, and MedChain~\citep{liu2024medchain} simulate multiturn patient-clinician interactions, while MMedAgent-RL~\citep{xia2025mmedagent}, MedAgentBoard~\citep{zhu2025medagentboard}, and the MAD framework~\citep{smit2023are} explore multi-agent training and collaboration strategies, finding that multi-agents do not always outperform strong LMM. 

Beyond text-based clinical workflows, broad reviews trace the evolution of generative AI from pure text LLMs to multimodal medical systems~\citep{buess2025large}. Finally, some studies address evaluation gaps and modality-specific challenges, such as MedHallu~\citep{pandit2025medhallu}, which focuses on hallucination detection, and work by~\citep{thapa2025disentangling} that disentangles knowledge from reasoning to expose benchmark inflation. In vision-language domains, PMC-VQA~\citep{zhang2024pmc} advances medical VQA through visual instruction tuning, MedRepBench~\citep{shang2025medrepbench} evaluates vision-language models to interpret complex medical reports, and Med3DInsight~\citep{chen2025enhancing} improves 3D image understanding by leveraging 2D LMM pretraining. In computational and digital pathology, studies emphasize clinical practice integration~\citep{hijazi2024digital} and efficient multimodal reasoning through rationale discovery and dynamic token allocation~\citep{xu2025discovering}.
In contrast, our work centers on multi-step and in-depth explorative insight discovery, pushing the boundaries of traditional evaluations to uncover deeper, more nuanced insights.

\subsection{Data Insight Agents and Benchmarks}

Works on LLM-driven data analysis have produced benchmarks, datasets, and agentic frameworks that move beyond single-query answers to multi-step analytical workflows. Text-to-SQL efforts include FinSQL~\citep{zhang2024finsql}, Spider 2.0~\citep{leispider}, EHRSQL~\citep{lee2022ehrsql}, and PRACTIQ~\citep{dong2025practiq}, which address domain-specific querying, complex multi-step SQL, and conversational ambiguity. To benchmark interactive program synthesis in computational notebooks, ARCADE~\citep{yin2023natural} evaluates multi-turn natural language to Python/Pandas code generation by capturing rich context and execution states. For visualization, VisEval~\citep{chen2024viseval} provides a comprehensive evaluation framework, while MatPlotAgent~\citep{yang2024matplotagent} and nvAgent~\citep{ouyang2025nvagent} propose multi-agent workflows to iteratively generate and validate visualizations, demonstrating significant improvements. In addition, InfiAgent-DABench~\citep{hu2024infiagent} introduces a broad benchmark for assessing LLM-based data analysis agents, and DAgent~\citep{xu2025dagent} extends this by generating complete analytical reports from relational databases. Other works target end-to-end insight generation. For example, InsightBench~\citep{sahuinsightbench}, InsightPilot~\citep{ma2023insightpilot}, InsightLens~\citep{weng2025insightlens}, and an LLM-based SQL decomposition approach~\citep{perez2025llm} cover multi-step discovery, autonomous exploration, and insight organization. Our MedInsightBench and MedInsightAgent constitute the first comprehensive and high-quality benchmark and the first automated and efficient multi-agent framework for medical insight discovery, respectively.

\section{Dataset Analysis}
\label{appendix:dataset}
\subsection{Detailed statistics of MedInsightBench}
\begin{figure}[htbp!]
    \centering
    \includegraphics[width=\linewidth]{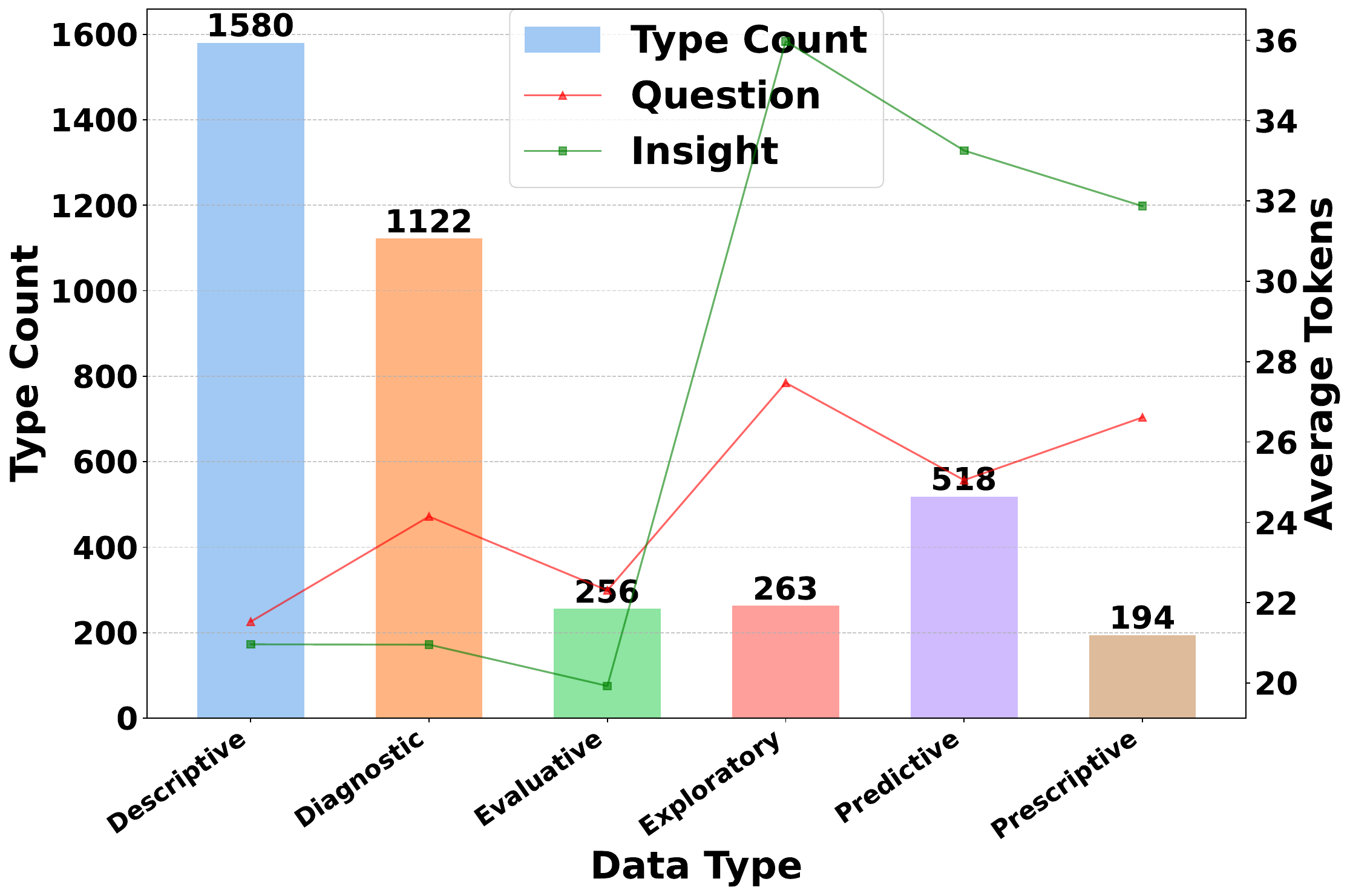}
    \caption{The distribution of different Insight Types in MedInsightBench and the average token count of Question and Insight in each type of data.}
    \label{fig:stat_left}
\end{figure}

\begin{figure}[htbp!]
    \centering
    \includegraphics[width=\linewidth]{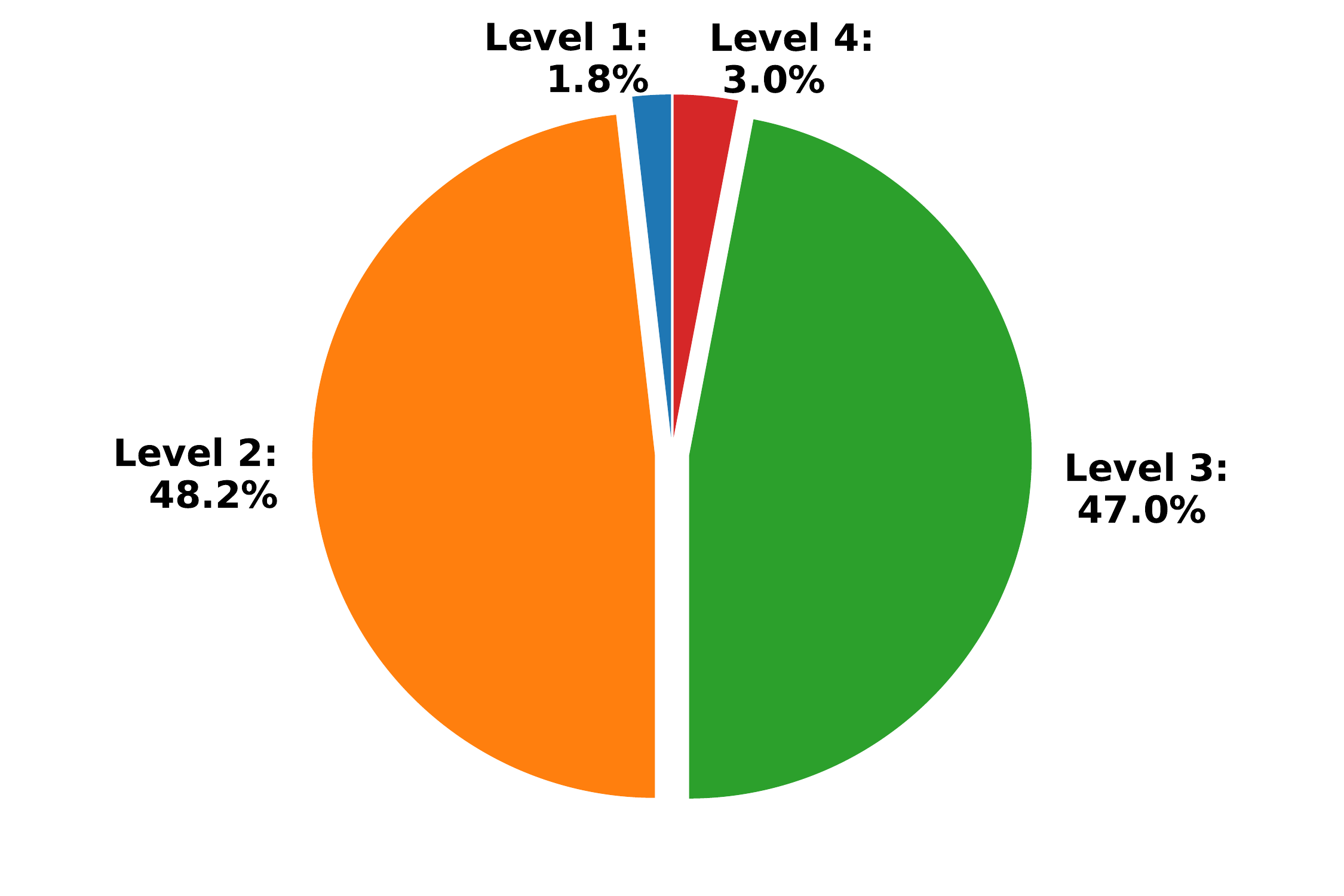}
    \caption{The distribution of different difficulty levels in MedInsightBench.}
    \label{fig:stat_right}
\end{figure}

Figure \ref{fig:stat_left} and Figure \ref{fig:stat_right} present detailed statistical information on MedInsightBench, including the distribution of different Insight categories, the average number of tokens in Questions and Insights per category, and the distribution across different difficulty levels.

\subsection{Insight Types}
\label{sec:insight_types}
In this paper, we provide a comprehensive interpretation of data insights by six insight categories. A detailed description of each insight category is provided below:

\begin{itemize}
\item \textbf{Descriptive}: In the medical context, descriptive insights summarize what has already occurred by aggregating and visualizing historical clinical and operational data. For example, charts of monthly inpatient admissions by diagnosis, trends in laboratory test volumes, or distributions of medication use between departments, so clinicians and administrators can quickly understand the current and past state of patients and services.
\item \textbf{Diagnostic}: Diagnostic insights explain why the observed clinical or operational patterns occurred by identifying correlations, temporal associations, and plausible causal factors, such as linking a rise in postoperative infections to a change in sterilization procedures, a particular implant type, or changes in staffing, helping teams prioritize investigations and corrective actions.
\item \textbf{Predictive}: Predictive insights use historical patient records, longitudinal vitals, laboratory trajectories, imaging characteristics, and social determinants to forecast future outcomes or events, such as 30-day readmission risk, likelihood of ICU transfer, or expected lab deterioration, providing probabilities and confidence estimates to inform proactive clinical planning.
\item \textbf{Prescriptive}: Prescriptive insights translate predictions and diagnostics into concrete, actionable recommendations that balance benefits, risks, and constraints, for example, suggesting personalized treatment adjustments, targeted follow-up schedules, or resource allocation strategies (e.g., bed assignment or staffing changes) designed to improve results or operational efficiency.
\item \textbf{Evaluative}: Evaluative insights assess the quality, reliability, and limitations of the data and analyze themselves by auditing data completeness, bias, model calibration, and external validity, for example, reporting subgroup performance disparities of a mortality model or highlighting key missing variables that undermine the conclusions.
\item \textbf{Exploratory}: Exploratory insights search for unknown or unexpected patterns without a prior hypothesis, using techniques such as clustering, anomaly detection, and dimensionality reduction to uncover new patient subgroups, unusual temporal events, or latent relationships, such as discovering a previously unrecognized phenotype associated with distinct biomarker patterns that merits further clinical study.
\end{itemize}

\subsection{Difficulty Level Definition}

The difficulty level of each sample is determined using four complementary dimensions:
(i) visual complexity of the pathology image,
including tissue heterogeneity and morphological ambiguity;
(ii) report complexity, such as terminology density and the number of pathological findings;
(iii) cross-evidence dependency, reflecting whether insights require integrating multiple evidence snippets or image regions; and
(iv) reasoning depth, measuring the extent of multi-step clinical inference and synthesis required for meaningful insight discovery.
Each sample is assigned a difficulty level from 1 to 4 through LLM-assisted estimation followed by human verification.

\section{Details of Agentic methods and Implementation}
\label{appendix:details_imple}
\subsection{Chain-of-Thought (CoT)}

Chain-of-Thought~\citep{wei2022chain} prompting encourages the model to produce explicit intermediate reasoning steps before giving the final answer. This simple prompting strategy has been shown to improve multi-step reasoning by making latent inference more explicit and structured.

For each sample, we provide the pathology image and the analytical goal to the backbone LMM and prompt it to reason step by step about salient pathological findings, relevant evidence, and the final medical insight. The model is instructed to first describe the most relevant visual cues, then infer the clinically meaningful implication of those cues, and finally produce a concise insight aligned with the goal. CoT does not use external tools or multi-turn interaction; it serves as a reasoning-only baseline that isolates the effect of explicit intermediate reasoning from tool use or orchestration.

\subsection{ReAct}

ReAct~\citep{yao2023react} interleaves reasoning traces with task-specific actions, allowing the model to alternate between thinking and acting. The key idea is that reasoning helps the agent track and update its plan, while actions allow it to gather external information from tools or environments.

We implemented a ReAct structure agent and equipped it with general tools, including a computation module and a web-search interface. For comparison, we adapted ReAct to the same pathology-insight setting by letting the model alternate between thought steps and tool calls. At each step, the agent may inspect the image, request external retrieval, or perform lightweight computation before producing the next reasoning step, following the broader trend toward active and selective visual reasoning in multimodal agents~\citep{bai2026glanceorgazeincentivizinglmmsadaptively}.

\subsection{Plan-and-Solve}

Plan-and-Solve~\citep{wang2023plan} prompting first asks the model to produce a decomposition of the task into smaller subtasks and then execute them in order. This design targets missing-step errors in zero-shot reasoning and encourages a more complete problem-solving trajectory.

For each case, the model first generates a short plan, such as: (1) identify salient morphological patterns, (2) map them to the analytical goal, (3) verify whether the evidence supports the target insight, and (4) summarize the conclusion. The model then executes the plan sequentially and outputs the final insight set. Unlike ReAct, Plan-and-Solve does not require an explicit action-feedback loop; instead, it acts as a planning-first baseline to test whether structured decomposition alone can improve insight discovery.

\subsection{Self-Refine}

Self-Refine~\citep{madaan2023selfrefine} improves an initial output through iterative self-feedback and revision. The same model first produces a draft, then critiques it, and finally refines the draft based on its own feedback. The method is training-free and uses the same LLM as generator, feedback provider, and refiner.

We prompt the model to first generate an initial insight draft from the pathology image and goal. Then we ask it to critique the draft along several criteria: goal relevance, grounding in visible evidence, completeness, and redundancy. After the critique, the model is prompted to revise the draft once or twice, keeping only clinically supported statements. This baseline tests whether test-time self-editing alone can close the gap to agentic orchestration.

\subsection{Reflexion}

Reflexion~\citep{shinn2023reflexion} reinforces language agents through verbal feedback rather than weight updates. After each attempt, the agent produces a reflective textual memory from task feedback and reuses that memory to improve later decisions in subsequent trials.

For each sample, the agent first generates an initial analysis and then writes a short reflection summarizing what was missing, what may have been unsupported, and what additional evidence should be checked. This reflection is stored in a case-level memory buffer and is fed back into the next attempt. Because MedInsightBench is evaluated case by case, the memory is reset between samples to avoid cross-case leakage. This baseline isolates the value of verbal self-correction and persistent memory in medical insight discovery.

\section{Evaluation Design and Model Agnosticism}
A common concern in evaluating agent-based benchmarks is the potential coupling between the data structure and the proposed agent's pipeline. To address this, we emphasize that MedInsightBench is designed to be model-agnostic. The benchmark evaluates the quality of the derived insights rather than the specific internal mechanism of the agent.

As demonstrated in our experiments (Table~\ref{tab:main_results}), we evaluate a spectrum of paradigms:
\begin{enumerate}
    \item Direct Decoding (e.g., GPT-4o, Deepseek-VL2), which relies on single-turn inference;
    \item General Agent Frameworks (e.g., ReAct), which utilize iterative prompting;
    \item MedInsightAgent, our specialized pipeline.
\end{enumerate}

The ability of MedInsightBench to distinguish performance across these distinct paradigms proves that the benchmark is not tailored to a single agent architecture but effectively assesses the inherent medical reasoning capability of the underlying LMMs. This confirms that MedInsightBench serves as an objective testbed for various clinical reasoning workflows.

\section{More Experimental Results}
\label{appendix:experiment}

\begin{figure}[htbp!]
    \centering
    \includegraphics[width=\linewidth]{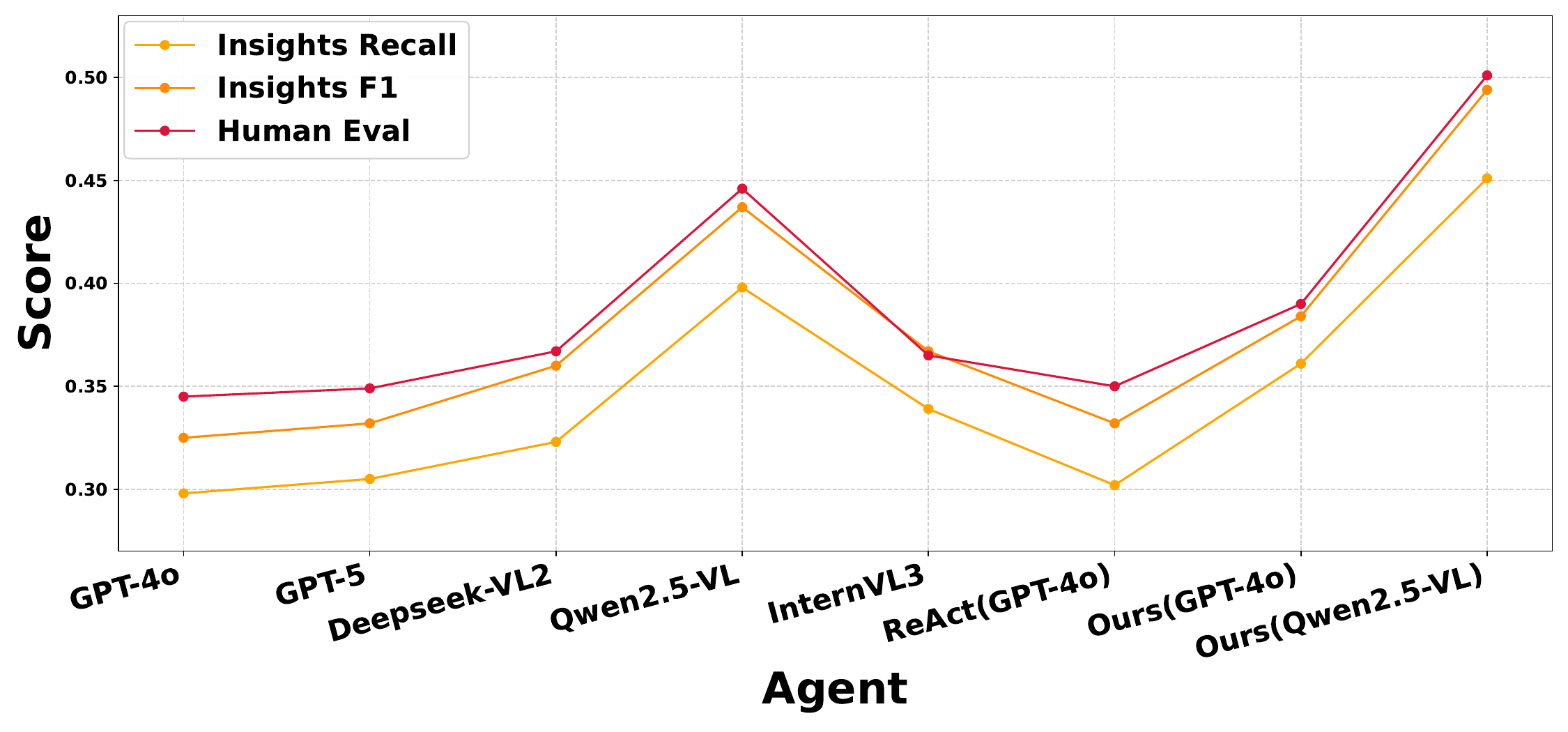}
    \caption{Comparison of G-Eval scores in Insight Recall, Insight $F1$, and Expert Scores in Human Evaluation.}
    \label{fig:human_vs_f1}
\end{figure}

\paragraph{Insight $F1$ Score as a Reliable Proxy for Expert-level Evaluation.}
To validate the reliability of our automated metrics, we conducted a validation study comparing automated scores with expert human evaluations. As shown in Figure~\ref{fig:human_vs_f1}, Insight $F1$ consistently demonstrates a higher correlation with human scores compared to Insight Recall, which tends to overestimate performance. This alignment confirms that Insight $F1$ serves as a more faithful proxy for assessing complex medical-insight discovery capabilities. Furthermore, our manual audit of predicted insights reveals that while automated metrics provide a scalable evaluation, they occasionally overestimate the novelty of insights. Consequently, we implemented an additional round of manual verification on top-performing models to ensure that the reported Novelty scores rigorously reflect clinical significance rather than superficial token-level variation.

\paragraph{Statistical Reliability of Fine-Grained Comparisons}
To assess the statistical reliability of the fine-grained comparisons, we conducted a bootstrap significance test with $B=10{,}000$ resamples on the primary results. As shown in Table~\ref{tab:bootstrap_significance}, MedInsightAgent achieves the highest G-Eval $F1$ of $0.5242 \pm 0.0090$, with a relatively narrow 95\% confidence interval. Its improvements over CoT, Qwen3-VL-32B, and GPT-4o are highly significant ($p<0.001$), while the gains over Reflexion and ReAct with the same tools are also significant at the $\alpha=0.05$ level ($p=0.011$ and $0.010$, respectively). These results provide statistical evidence that the performance advantage of MedInsightAgent is consistent across resamples rather than attributable to random variation.

\begin{table}[htbp!]
\centering
\resizebox{\linewidth}{!}{
\begin{tabular}{lcc}
\toprule
\textbf{Comparison} & \textbf{G-Eval $F1$ (Mean $\pm$ 95\% CI)} & \textbf{$p$-value} \\
\midrule
MedInsightAgent & $0.5242 \pm 0.0090$ & -- \\
ReAct + same tools & $0.5067 \pm 0.0096$ & $0.010$ \\
Reflexion & $0.5063 \pm 0.0097$ & $0.011$ \\
CoT & $0.4826 \pm 0.0105$ & $<0.001$ \\
Qwen3-VL-32B & $0.4774 \pm 0.0109$ & $<0.001$ \\
GPT-4o & $0.3250 \pm 0.0138$ & $<0.001$ \\
\bottomrule
\end{tabular}
}
\caption{Bootstrap significance test on the primary results ($B=10{,}000$). G-Eval $F1$ is reported as mean $\pm$ 95\% confidence interval. $p$-values are computed against MedInsightAgent.}
\label{tab:bootstrap_significance}
\end{table}

\begin{table*}[htbp]
\fontsize{9.0pt}{9.0pt}\selectfont 
\renewcommand\arraystretch{1.2}
\setlength{\tabcolsep}{2mm}
\centering
\begin{tabular}{lccccccc}
\toprule
\textbf{Configuration} & \textbf{ISM} & \textbf{WRM} & \textbf{IAT} & \textbf{FQC} & \textbf{G-Eval $F1$} & \textbf{Novelty} & $\boldsymbol{\Delta}$\textbf{$F1$} \\
\midrule
MedInsightAgent & \checkmark & \checkmark & \checkmark & \checkmark & \textbf{0.5242} & \textbf{0.4995} & -- \\
w/o ISM \& WRM & $\times$ & $\times$ & \checkmark & \checkmark & 0.4996 & 0.4607 & $-0.0246$ \\
w/o ISM \& WRM \& IAT & $\times$ & $\times$ & $\times$ & \checkmark & 0.4889 & 0.4524 & $-0.0353$ \\
\bottomrule
\end{tabular}
\caption{Joint ablation results for analyzing interactions among the specialized tools in MedInsightAgent.}
\label{tab:joint_ablation}
\end{table*}

\paragraph{Joint Ablation Reveals Complementary Component Interactions}
To further examine component interactions, we conduct joint ablation experiments beyond the individual tool-level ablations in Table~\ref{tab:ablation_study_results}. As shown in Table~\ref{tab:joint_ablation}, removing both ISM and WRM reduces G-Eval $F1$ from 0.5242 to 0.4996, a drop of 0.0246, slightly exceeding the additive prediction of 0.0237 (=0.0116+0.0121). This suggests a positive interaction between ISM and WRM, where image summaries provide visual-semantic anchors that improve the effectiveness of retrieval within VRF. Removing ISM, WRM, and IAT further reduces $F1$ to 0.4889, corresponding to a total loss of 0.0353 relative to the full model. Notably, this zero-tool configuration still outperforms CoT (0.4826), indicating that the structured VRF$\rightarrow$AIA$\rightarrow$FQC pipeline provides standalone value beyond prompting-based reasoning. Overall, the results suggest that MedInsightAgent benefits from both its structured pipeline and the complementary effects of its specialized tools.

\begin{table*}[t!]
\centering
\fontsize{9pt}{9pt}\selectfont 
\resizebox{\textwidth}{!}{
\begin{tabular}{p{0.28\textwidth} p{0.30\textwidth} p{0.34\textwidth}}

\toprule
\textbf{Ground-Truth Insight} & \textbf{Output of GPT-4o} & \textbf{Output of MedInsightAgent(GPT-4o)} \\

\midrule
\rowcolor{gray!30} 
\multicolumn{3}{c}{\textbf{\textit{Case 139}:} TCGA-EJ-A7NM with Difficulty \textbf{Level 2}} \\
\midrule
Lymphovascular invasion and extensive perineural invasion suggest increased metastatic potential; consider systemic therapy evaluation. &
\underline{Absence of perineural invasion} in the visible sections may impact staging. & 
\textbf{Perineural invasion suggests a more aggressive tumor}, which might \textbf{increase the likelihood of cancer recurrence} and affect treatment decisions. \\
\midrule
Multifocal extracapsular extension and seminal vesicle invasion indicate locally advanced disease with higher recurrence risk; discuss adjuvant therapy options. &
Stromal desmoplasia is present, indicating \underline{possible tumor-host interaction} and invasive behavior. &
\textbf{A strong stromal reaction, known as desmoplasia, often indicates a more aggressive tumor} behavior and higher grade in prostate cancer, potentially impacting treatment strategies and prognosis. \\

\midrule
\rowcolor{gray!30} 
\multicolumn{3}{c}{\textbf{\textit{Case 122}:} TCGA-IN-A7NT with Difficulty \textbf{Level 3}} \\
\midrule
Angiolymphatic invasion is present. &
\underline{No definitive lymphovascular invasion} observed, although detailed review and further sections may be warranted. & 
\textbf{The presence of lymphovascular invasion} can suggest a higher risk of metastasis, informing treatment decisions and prognosis. \\
\midrule
Poorly differentiated (G3) adenocarcinoma suggests more aggressive tumor biology; consider comprehensive treatment approach. &
There are poorly formed glands and irregular cell clustering, which \underline{may indicate} a higher histologic grade. &
The presence of \textbf{pronounced cellular atypia and high mitotic rate suggests a more aggressive tumor behavior}, potentially correlating with a higher risk of nodal metastasis and impacting therapeutic decisions. \\

\midrule
\rowcolor{gray!30} 
\multicolumn{3}{c}{\textbf{\textit{Case 13}:} TCGA-05-4250 with Difficulty ~~~\textbf{Level 4}} \\
\midrule
Poorly differentiated grade 3 tumor with lymph node metastases, vascular invasion, and R2 resection status indicates high risk of recurrence and poor prognosis; recommend multidisciplinary discussion for adjuvant therapy consideration. &
The presence of irregular nests may indicate an aggressive phenotype. & 
These pathological features indicate \textbf{aggressive tumor behavior}, which is crucial for \textbf{determining prognosis and guiding effective treatment strategies}. \\
\midrule
Multifocal invasion of blood vessels is identified. &
\underline{No clear lymphovascular invasion} is observed in the current section. &
\textbf{This pattern of tissue invasion suggests a higher risk of cancer spreading beyond its origin}, which could impact treatment strategies. \\

\bottomrule
\end{tabular}
}
\caption{Case study of insight discovery. We selected three cases across different difficulty levels, with the bolded statements highlighting instances where MedInsightAgent demonstrated superior performance and the underlined parts show the defects in the output of GPT-4o.}
\label{tab:case_study}
\end{table*}

\section{Case Study}
\label{appendix:case_study}

Table~\ref{tab:case_study} presents representative cases across multiple difficulty levels, revealing several systematic differences between direct GPT-4o decoding and MedInsightAgent (GPT-4o). GPT-4o frequently recognizes local pathological patterns but fails to infer their higher-level clinical significance, leading to incomplete, overly generic, or occasionally contradictory interpretations. Its outputs often remain descriptive rather than analytical, especially in cases requiring reasoning about prognosis, metastasis risk, or treatment implications.

In contrast, MedInsightAgent demonstrates a stronger ability to transform pathological evidence into clinically actionable insights. The agent more consistently links findings such as lymphovascular invasion, poor differentiation, and stromal desmoplasia with tumor aggressiveness, recurrence risk, and therapeutic relevance. This advantage becomes increasingly pronounced as case difficulty increases, where accurate interpretation requires integrating multiple pathological signals into a coherent medical assessment rather than describing isolated observations.

The examples also highlight an important limitation of current general-purpose LLMs on MedInsightBench: even when relevant visual or pathological cues are partially recognized, the models often lack the domain-specific reasoning needed to derive expert-level conclusions. Although MedInsightAgent substantially improves evidence alignment and clinical reasoning quality, some outputs remain relatively abstract, suggesting that further advances in medical grounding and pathology-specific knowledge are still necessary.

\section{Detailed Evaluation Framework}
\label{appendix:evaluation_details}

Medical insight discovery differs fundamentally from standard closed-form question answering, as multiple clinically meaningful interpretations may exist for a single pathological image and report. Recent studies similarly highlight the difficulty of evaluating multimodal reasoning, where errors may arise at different stages of the reasoning trajectory~\citep{zhou2026whathowunveilingprocess}. Related work further points to limitations of automated reward and evaluation systems in multimodal settings, motivating the integration of automatic assessment with human verification~\citep{chan-etal-2026-omni}. However, these conventional approaches suffer from two critical limitations. First, they often overlook unsupported or hallucinated generations, failing to penalize clinically invalid conclusions that may appear partially matched. This limitation motivates explicit verification mechanisms, consistent with prior work showing the value of step-wise verification and correction for improving the reliability of complex reasoning~\citep{shi2025legalreasoner}. Second, they lack the capability to recognize novel yet clinically significant insights beyond predefined annotations, which is vital for exploratory analysis.

Current benchmarks such as InsightBench~\citep{sahuinsightbench} exacerbate these issues by relying exclusively on automated text-match metrics or limited evaluators like LLAMA-3-Eval, which are susceptible to model biases. Similar concerns regarding the reliability and completeness of LLM evaluation have also been raised in safety-oriented benchmarking~\citep{cao2025safelawbench}. Furthermore, these evaluations focus solely on discovering pre-annotated insights, neglecting the identification of erroneous outputs and failing to reward the discovery of novel findings. To address these shortcomings, drawing inspiration from InsightEval~\citep{zhu-etal-2026-insighteval}, we adopt a comprehensive, four-dimensional automated evaluation framework designed to rigorously measure the medical insight discovery capabilities of Large Multimodal Models (LMMs) and Agents. By comparing generated insights ($I$) with annotated ground-truth insights ($GT$), our framework employs four principal measures: Insight Recall, Insight Precision, Insight $F1$, and Insight Novelty. These metrics jointly evaluate coverage, faithfulness, balanced analytical quality, and exploratory capability, providing a robust assessment that captures both the reliability and the open-ended discovery potential of advanced medical AI systems.

\subsection{Insights Recall Evaluation}

To assess if ground-truth insights are discovered, we need to calculate the recall rate by adapting the iterative matching protocol. Insight Recall measures whether a model successfully recovers the key evidence-backed insights that should be discovered from the pathology image and associated report. In the clinical setting, high recall indicates strong coverage of diagnostically important findings and reduced risk of missing clinically relevant evidence. We count the scores between each ground-truth insight ($gt \in GT$) and each generated insight ($i \in I$). Then we record the highest-scoring counterpart based on each ground-truth insight ($gt$) and calculate the expectation score ($E$) as the final output. The formula for recall evaluation is shown as in Eq.~\ref{recall_equation}, with $S$ representing the evaluator, such as ROUGE-1 or G-Eval.

\begin{equation}
\!\!\!\mathrm{Score_{recall}}(\mathcal{S})
\!\!\!\;=\;\!\!\!
E_{gt \sim \mathrm{Unif}(GT)}
\bigl[\max_{i \in I}\,\mathcal{S}(gt,i)\bigr]
\label{recall_equation}
\end{equation}

\subsection{Insights Precision Evaluation}

Only focusing on the recall rate may overlook the possibility that agents generate irrelevant or unnecessary insights. To address this limitation, it is essential to further evaluate the accuracy of each generated insight to enhance the overall evaluation system. Insight Precision evaluates whether generated insights are clinically grounded, factually consistent with the available evidence, and free from unsupported extrapolation. In medical applications, precision is particularly important because plausible but incorrect analytical conclusions may introduce substantial clinical risk. Similarly, we also enumerate the scores between the ground truth and the generated insight. To calculate the precision rate, we need to record the highest score based on each generated insight ($I$). The formula for precision evaluation is presented as in Eq.~\ref{precision_equation}.

\begin{equation}
\!\!\!\!\mathrm{Score_{precision}}(\mathcal{S})
\!\!\!\;=\;\!\!\!
E_{i \sim \mathrm{Unif}(I)}
\bigl[\max_{gt \in GT}\,\mathcal{S}(i,gt)\bigr]
\label{precision_equation}
\end{equation}

\subsection{Insights $F1$ Evaluation}

To comprehensively assess the capability of insight discovery, we proposed an insight $F1$ score. Insight $F1$ balances Recall and Precision, jointly measuring analytical completeness and clinical faithfulness. Compared with Recall alone, $F1$ penalizes unsupported or hallucinated outputs while still rewarding broad evidence coverage. Compared with Precision alone, it avoids overly conservative systems that generate only a small subset of safe insights. Therefore, we treat Insight $F1$ as the primary evaluation metric for pathology-grounded medical insight discovery. With the recall score and the precision score, we can calculate the insight $F1$ score through the formula in Eq.~\ref{F1_equation}.

\begin{equation}
\!\!\!\mathrm{Score_{F1}}
\!\!\!\;=\;\!\!\!
\frac{2 * Score_{recall} * Score_{precision}}{Score_{recall}+Score_{precision}}
\label{F1_equation}
\end{equation}

\subsection{Insights Novelty Evaluation}

Medical insight discovery is inherently exploratory; beyond merely recovering annotated findings, a robust analytical system should identify additional clinically meaningful observations that remain consistent with available evidence. To address the limitations of evaluating models solely through alignment with ground-truth ($GT$) insights, it is essential to quantify the capacity for discovering novel insights.

We define Insight Novelty as the ability of a model to generate valid insights that extend beyond the reference annotations. Importantly, this evaluation is predicated on the principle that novelty is only meaningful when the generated insight is evidence-consistent and clinically plausible. To avoid rewarding unsupported generations (hallucinations), novelty is evaluated exclusively after a rigorous faithfulness verification: only insights that have been successfully filtered or validated during the precision evaluation pipeline are considered for novelty assessment.

For insights that are not explicitly matched to $GT$ references, we perform a secondary evaluation focused on innovation. During this process, we utilize three distinct LMMs to mitigate individual model bias, employing a Chain-of-Thought (CoT) reasoning framework to enhance judgment accuracy. To ensure the evaluators maintain a comprehensive clinical perspective, we provide them with multi-modal inputs, including the clinical goal, the medical image, and historical insights. An insight is classified as novel when at least two out of the three models verify its correctness.

The formula for novelty evaluation is expressed as in Eq.~\ref{novelty_equation}, where $\mathrm{LMM}_j(i)\in\{0,1\},\quad \delta\in\{0,1\}$ denotes the correctness judgment of the $j$-th evaluator, $\mathbf{1}$ is the indicator function, $M$ and $N$ represents the total number of candidate insights identified as matching and non-matching (and thus potentially novel) during the precision evaluation phase, respectively.

\begin{equation}
    \mathrm{Score_{novelty}} = \frac{M + \delta \cdot C}{N + M}
    \label{novelty_equation}
\end{equation}
where $C$ is the count of novel samples:
\begin{equation}
    C = \sum_{i=1}^{N} \mathbf{1} \Bigl( \sum_{j=1}^{3} \mathrm{LLM}_j(i) \ge 2 \Bigr)
\end{equation}

When $\delta=1$, the formula calculates the Innovation score. For comparison, we set $\delta=0$ to obtain the Original score during the evaluation.

\section{Algorithm of MedInsightAgent}
\label{appendix:algorithm}
The general algorithm framework of MedInsightAgent is shown in Algorithm \ref{alg:framework}.

\begin{algorithm*}[t!]
\caption{Overall Multi-Agent Insight Mining Framework}
\label{alg:framework}

\begin{algorithmic}[1]

\Require Medical image $I$, analysis goal $G$
\Ensure Final insights $S$, answers $A$, root questions $Q$, follow-up questions $T$

\vspace{0.3em}

\State \textbf{Initialize:}
$\;F, K, D, Q, A, S, T \gets \emptyset$

\vspace{0.5em}

\State \textbf{Stage 1: Visual Root Finder (VRF)} \hfill \textit{Eq.~(1)}

\State Extract visual summary and keywords:
\State \makebox[\linewidth][c]{$(F, K) \gets \mathcal{IS}_M(I)$}

\State Retrieve external domain knowledge:
\State \makebox[\linewidth][c]{$D \gets \mathcal{WR}_M(K)$}

\State Generate root questions:
\State \makebox[\linewidth][c]{$Q \gets \mathcal{L}(I,G,F,D)$}

\vspace{0.5em}

\State \textbf{Stage 2: Analytical Insight Agent (AIA)} \hfill \textit{Eq.~(2)}

\For{each root question $q_i \in Q$}

    \State Extract question-specific evidence:
    \State \makebox[\linewidth][c]{$E_i \gets \mathcal{LAT}(I,q_i)$}

    \State Generate answer and insight:
    \State \makebox[\linewidth][c]{$(a_i,s_i) \gets \mathcal{G}(q_i,I,E_i)$}

\EndFor

\State Aggregate outputs:
\State \makebox[\linewidth][c]{$A \gets \{a_i\}_{i=1}^{m}, \quad S \gets \{s_i\}_{i=1}^{m}$}

\vspace{0.5em}

\State \textbf{Stage 3: Follow-up Question Composer (FQC)} \hfill \textit{Eq.~(3)}

\State Generate candidate follow-up questions:
\State \makebox[\linewidth][c]{$C \gets \mathcal{F}(I,G,A,F,D,Q)$}

\State Select optimal follow-up question:
\State \makebox[\linewidth][c]{$c_{\text{best}} \gets \mathcal{S}(C)$}

\State Update follow-up set:
\State \makebox[\linewidth][c]{$T \gets T \cup \{c_{\text{best}}\}$}

\vspace{0.5em}

\State \textbf{Iterative Refinement}

\While{stopping criterion not met}

    \State Feed $c_{\text{best}}$ back to AIA
    \State Update $A$, $S$, and $Q$

\EndWhile

\vspace{0.3em}

\State \Return $(S,A,Q,T)$

\end{algorithmic}
\end{algorithm*}

\section{Detailed Human Annotation and Verification Protocol}
\label{appendix:human_annotation_details}

To ensure the statistical rigor, medical accuracy, and clinical relevance of the Medical Insight Benchmark, we implemented a comprehensive human-in-the-loop verification and quality control framework throughout our data construction pipeline.

\subsection{Annotator Cohort and Inter-Annotator Agreement}
Our primary data annotation and verification tasks were executed by a dedicated cohort of domain-specific annotators consisting of students with backgrounds in pathology, medical image analysis, or clinical data interpretation. All annotators were provided with detailed scoring guidelines and standardized evaluation criteria before annotation. To guarantee the reliability and reproducibility of the manual curation process, we conducted a rigorous inter-annotator agreement analysis. 

We randomly sampled 100 independent cases across the dataset to undergo fully overlapped, blind double-annotation for both dataset quality checks and model evaluation scoring. We then computed three standardized statistical agreement metrics across raters:
\begin{itemize}[leftmargin=*,noitemsep,nolistsep]
\item \textbf{Intraclass Correlation Coefficient (ICC):} The average ICC across raters was approximately \textbf{0.82}, reflecting excellent reliability and strong numerical alignment in quantitative judgments.
\item \textbf{Krippendorff’s $\alpha$:} For categorical and qualitative assessments, Krippendorff’s $\alpha$ reached approximately \textbf{0.84}, demonstrating robust consensus that exceeds the standard academic threshold (0.80) for reliable data annotation.
\item \textbf{Pearson Correlation Coefficient:} The pairwise linear correlation yielded an average coefficient of \textbf{0.76}, further substantiating the high consistency and systematic agreement across individual human annotators.
\end{itemize}

\subsection{Granular Verification Protocols across Pipeline Stages}
The human verification was not a superficial passive review, but rather a highly structured process governed by explicit checklists and clinical constraints tailored to each pipeline stage.

\paragraph{1. Image and Report Quality Assessment Checklist (Step 1 \& Step 2)}
For the pathological whole-slide images and their corresponding raw OCR report texts, annotators performed binary (0--1) correctness judgments and manual filtering based on a rigorous multi-tier checklist:
\begin{itemize}[leftmargin=*,noitemsep,nolistsep]
\item \textbf{Critical Clinical Checks (Mandatory Pass):} Annotators verified the explicit and unambiguous presence of the final clinical diagnosis, precise tumor size, histological grade/stage, lymph node status (node-positive vs. node-negative), surgical margin status, and primary anatomical site/laterality. Any sample with severe OCR corruption, unreadable text, or forbidden patient linking identifiers was immediately discarded.
\item \textbf{Helpful Auxiliary Checks:} Annotators cross-referenced the consistency of immunohistochemistry (IHC) or molecular marker results, documented patient clinical history, and proactively scanned for any internal contradictions within the text blocks.
\end{itemize}

\paragraph{2. Triplet Consistency and Logic Verification (Step 2 - Stage 4)}
To guarantee the logical integrity of the core benchmark components, annotators evaluated the generated \textit{question--insight--evidence} triplets. Rather than evaluating components in isolation, human annotators reviewed the triplets as unified cognitive chains, verifying that:
\begin{itemize}[leftmargin=*,noitemsep,nolistsep]
\item The formulated analytical question is fully answerable by, and directly grounded in, the provided clinical evidence excerpt.
\item The extracted insight remains strictly faithful to the original pathology report text, containing zero clinical hallucinations.
\item The insight successfully aligns with the overarching goal without exhibiting any over-extrapolation, unwarranted medical speculation, or logical leaps.
\end{itemize}

\paragraph{3. Post-Hoc Bias Elimination in Goal Generation (Step 3)}
To ensure that the analytical goals serve as valid priori guides rather than arbitrary post-hoc synthesis artifacts, the goals were strictly constrained. Annotators verified that each goal was systematically abstracted from the shared analytical trajectories of the underlying question lists. Guided by predefined evaluation prompts, a valid goal was required to: (i) capture a cohesive clinical direction; (ii) be highly actionable at an advanced diagnostic level; (iii) reflect clear clinical relevance; and (iv) strictly avoid introducing specific downstream implementation details or answers.

\subsection{Human Scoring Protocol for Benchmark Results}
In addition to dataset construction verification, human experts were deployed to provide annotation-based evaluation scores for validating baseline model outputs (as presented in Appendix~\ref{appendix:experiment}). 

Adhering to a standardized protocol inspired by G-Eval, annotators compared model- or agent-generated insights directly against our curated ground-truth insights. Human annotators evaluated the generated insights across multiple axes, including fine-grained semantic similarity, clinical accuracy, and reasoning plausibility. Each metric was quantified using an explicit 1--10 Likert scale, backed by specific anchoring definitions for each score tier to minimize subjective rating drift.

\subsection{Alignment Between Human and Automated Evaluation}

We further analyze the correlation between automated evaluation metrics and human judgments. Among all metrics, Insight $F1$ exhibits the strongest alignment with human evaluation, suggesting that balanced measurement of coverage and faithfulness better reflects clinically meaningful analytical quality than Recall alone.

In addition, although automated novelty scores are generally more optimistic in magnitude, they preserve the relative ranking of methods and remain well aligned with human assessments, supporting the validity of the proposed novelty evaluation protocol.

\begin{table*}[htbp!]
\centering
\fontsize{9.0pt}{9.0pt}\selectfont 
\renewcommand\arraystretch{1.2}
\setlength{\tabcolsep}{2mm}
\begin{tabular}{lccccc}
    \toprule
    \textbf{Model} & \textbf{Reviewed} &
    \textbf{Category A} & \textbf{Category B} &
    \textbf{Category C} & \textbf{Human Approval Rate} \\
    \midrule
    MedInsightAgent & 30 & 13 (43.3\%) & 13 (43.3\%) & 4 (13.4\%) & \textbf{26 (86.6\%)} \\
    Qwen3-VL-32B   & 30 & 8 (26.7\%)  & 13 (43.3\%) & 9 (30.0\%)  & 21 (70.0\%) \\
    Reflexion      & 30 & 9 (30.0\%)  & 13 (43.3\%) & 8 (26.7\%)  & 22 (73.3\%) \\
    \bottomrule
\end{tabular}
\caption{Supplementary human audit of insights that passed the Novelty filter. Category A denotes true clinical novelty, Category B denotes clinically relevant but already known information, and Category C denotes trivial or clinically uninformative statements. The human approval rate is the proportion of insights categorized as A or B.}
\label{tab:human_novelty_audit}
\end{table*}

\subsection{Human Audit of Clinically Novel Insights}
\label{app:human_novelty_audit}

Beyond the double-blind annotation study described above, we conducted a supplementary human audit to further assess the clinical validity of the insights identified by our Novelty metric. Specifically, we randomly sampled 30 Novelty-filtered insights from each of three representative methods and categorized them into three levels: \textbf{Category A}, true clinical novelty; \textbf{Category B}, clinically relevant but already known information; and \textbf{Category C}, trivial or clinically uninformative statements. As shown in Table~\ref{tab:human_novelty_audit}, MedInsightAgent achieves the highest proportion of genuinely novel clinical insights, with 43.3\% of audited insights assigned to Category A, compared with 26.7\% for Qwen3-VL-32B and 30.0\% for Reflexion. Meanwhile, only 13.4\% of MedInsightAgent's insights are categorized as trivial statements, substantially lower than Qwen3-VL-32B (30.0\%) and Reflexion (26.7\%). Overall, 86.6\% of MedInsightAgent's insights are judged by human evaluation to be either genuinely novel or clinically relevant (Categories A+B), compared with 70.0\% and 73.3\% for Qwen3-VL-32B and Reflexion, respectively. These results provide complementary evidence that our Novelty metric captures clinically meaningful discoveries rather than merely surface-level linguistic variations, and that the superior performance of MedInsightAgent is also reflected in human assessment of clinical value. While large-scale independent validation by practicing pathologists remains the gold standard and an important direction for future work, the combination of double-blind agreement analysis and targeted human auditing provides strong empirical support for the reliability and clinical validity of our automated evaluation framework.

\begin{table*}[htbp!]
\centering
\fontsize{9.0pt}{9.0pt}\selectfont 
\renewcommand\arraystretch{1.2}
\setlength{\tabcolsep}{2mm}
\resizebox{\textwidth}{!}{
\begin{tabular}{lccccc}
\toprule
\textbf{Method} & \textbf{API Calls / Sample} & \textbf{Avg. Input Tokens} & 
\textbf{Avg. Output Tokens} & \textbf{Wall-clock (s/Sample)} & 
\textbf{G-Eval $F1$} \\
\midrule
Direct Decoding & 1 & 2,352 & 521 & 15.8 & 0.4774 \\
CoT & 1 & 2,685 & 769 & 27.6 & 0.4826 \\
ReAct & 12.4 & 10,907 & 2,854 & 133.4 & 0.5067 \\
MedInsightAgent & 32 & 65,372 & 7,692 & 958.9 & \textbf{0.5242} \\
\bottomrule
\end{tabular}
}
\caption{Inference cost and performance comparison across different methods. 
All values are averaged per sample.}
\label{tab:computational_cost}
\end{table*}

\section{Inference Latency and Computational Cost Analysis}
\label{app:computational_cost}

\paragraph{MedInsightAgent achieves stronger performance with a substantial but interpretable inference overhead.}
To further analyze the inference efficiency of different approaches, we report their average API calls, input/output token consumption, wall-clock inference time, and G-Eval $F1$ per sample in Table~\ref{tab:computational_cost}. MedInsightAgent incurs substantially higher computational costs than Direct Decoding and CoT, with an average wall-clock time of 958.9 seconds per sample. Nevertheless, this overhead is accompanied by a consistent performance gain. Compared with ReAct, the closest agentic baseline, MedInsightAgent increases wall-clock inference time by approximately $7.2\times$, while improving G-Eval $F1$ from 0.5067 to 0.5242, corresponding to an absolute gain of 0.0175 and a relative improvement of approximately 3.5\%. Moreover, MedInsightAgent also achieves a more substantial improvement in Novelty, indicating that the additional computation contributes to discovering more diverse and non-trivial clinical insights. Given that MedInsightBench targets high-value clinical scenarios, such as tumor staging and biomarker discovery, where insight quality and clinical reliability are more important than minimizing latency alone, we consider this cost-performance trade-off reasonable. The relatively high computational cost is also largely attributable to deliberate design choices in the agent architecture rather than redundant computation. Specifically, the 32 API calls per sample are primarily induced by the ISM/WRM tools in VRF and the multi-step IAT iterations in AIA, where four follow-up questions are generated and processed through approximately eight steps each. In addition, the high input-token consumption is mainly caused by incorporating the complete image context into each IAT call, which is necessary to maintain full traceability and multimodal grounding when verifying each generated insight. Thus, the observed computational overhead reflects the explicit emphasis of MedInsightAgent on iterative clinical verification and faithful image-grounded reasoning, rather than an unintended inefficiency.

\section{The Use of Large Language Models (LLMs)}
We acknowledge the use of large language models (LLMs) as auxiliary tools in the preparation of this work, primarily in the following aspects:

\begin{enumerate}
    \item Dataset construction: During the dataset development process, we adopted an LLM-assisted approach combined with manual review. Specifically, LLMs were employed to refine and streamline the prompts used in data collection.
    \item Manuscript preparation: LLMs were utilized for word choice and grammar checking, as well as for polishing the language throughout the writing of this manuscript.
    \item The authors independently conceived and determined all research ideas, experimental designs, data analysis, and conclusions.
\end{enumerate}

\section{Examples case of MedInsightBench}
\label{appendix:example_case}
Tables \ref{data_example}, \ref{data_example_2}, and \ref{data_example_3} present Case 4 from MedInsightBench, which includes a specific goal, an image of medical cancer pathology, and a series of insights. Each insight consists of a question, an insight text, and an insight type.

\begin{table*}[htbp!]
\centering
\resizebox{\textwidth}{!}{
\begin{tabular}{p{0.25\textwidth} p{0.69\textwidth}}

\toprule
\textbf{Goal} & Correlate histopathologic features of the tongue carcinoma with staging parameters, margin status, nodal metastasis, and HPV status to guide prognostic assessment and treatment planning. \\

\midrule
\textbf{Pathological Image} & \includegraphics[width=10cm]{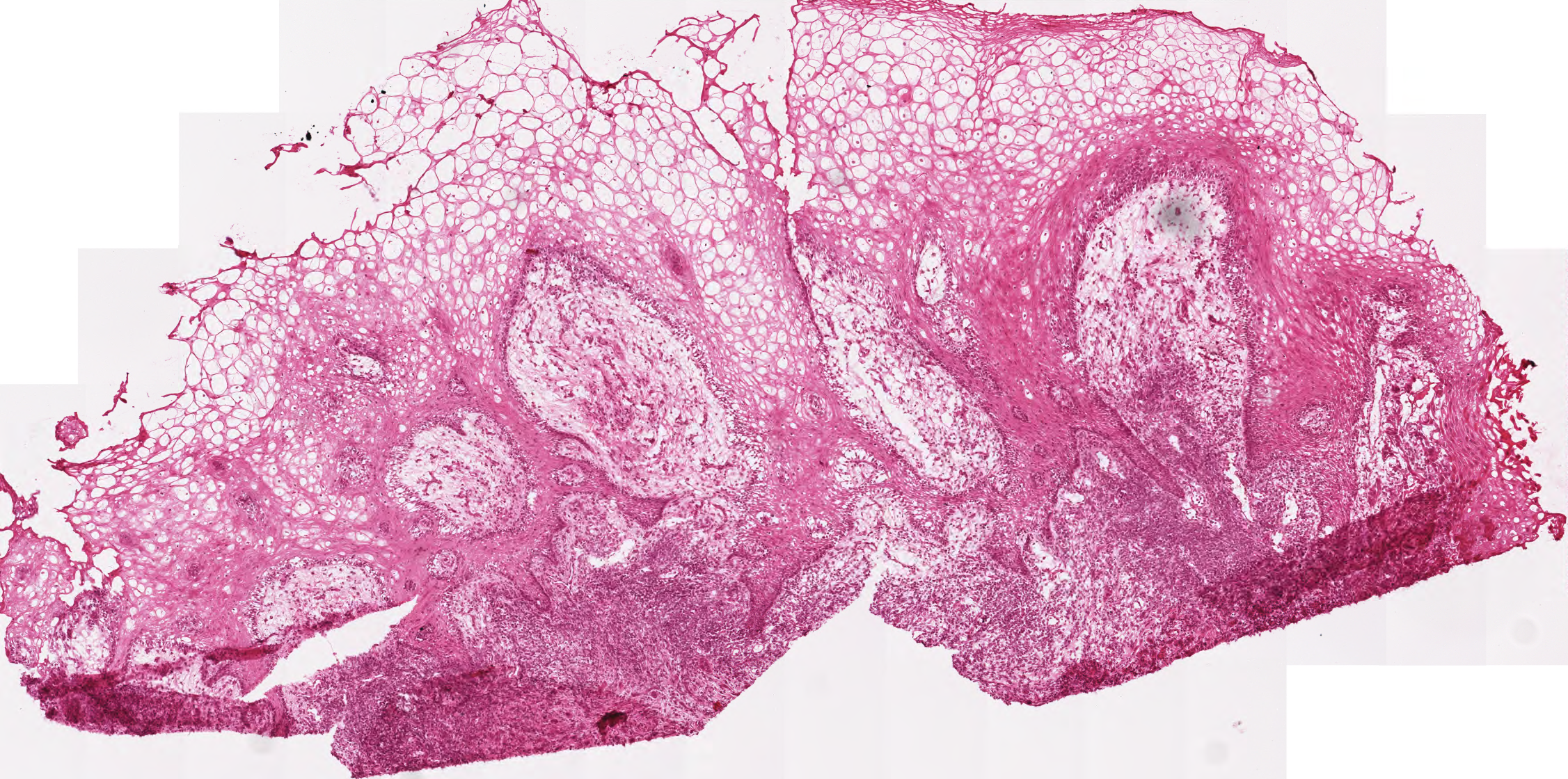} \\

\bottomrule
\end{tabular}
}
\caption{Goal and pathological image of Case 4 in MedInsightBench.}
\label{data_example}
\end{table*}

\begin{table*}[htbp!]
\centering
\resizebox{\textwidth}{!}{
\begin{tabular}{p{\textwidth}}
\toprule
\textbf{Insights Details}\\

\midrule
\textbf{Question 1}: What is the diagnosis, location, size, and depth of invasion as documented in the pathology report for the excised tongue specimen?

\textbf{Insight 1}: Invasive keratinizing squamous cell carcinoma identified in the right lateral tongue dorsum and lateral tongue, approximately 2.0 cm in greatest dimension with muscle invasion.

\textbf{Type 1}: Diagnostic \\
\midrule
\textbf{Question 2}: Does the pathology report confirm the presence of metastatic carcinoma in a lymph node from the right neck level 2, and what is the greatest dimension measurement of the identified tumor deposit?

\textbf{Insight 2}: Metastatic carcinoma identified in one lymph node from right neck level 2 with a tumor deposit measuring 1.9 cm.

\textbf{Type 2}: Diagnostic \\
\midrule
\textbf{Question 3}: What is the status of all surgical margins regarding tumor presence in the final pathology report?

\textbf{Insight 3}: Final surgical margins are negative for tumor in all specimens.

\textbf{Type 3}: Descriptive \\
\midrule
\textbf{Question 4}: What are the specific pathologic stage descriptors for the primary tumor, regional lymph nodes, and the number of lymph nodes examined versus involved as documented in the report?

\textbf{Insight 4}: Pathologic staging is pT1 pN1 with 44 lymph nodes examined and 1 involved.

\textbf{Type 4}: Descriptive \\
\midrule
\textbf{Question 5}: What are the reported findings regarding histologic grade, extracapsular extension, perineural invasion, and bony/cartilage invasion?

\textbf{Insight 5}: Tumor is moderately-differentiated squamous cell carcinoma without extracapsular extension, perineural invasion, or bony/cartilage invasion.

\textbf{Type 5}: Descriptive \\
\midrule
\textbf{Question 6}: What were the results of HPV testing for both p16 immunohistochemistry and high risk HPV in situ hybridization?

\textbf{Insight 6}: HPV testing performed shows p16 negative by immunohistochemistry and high risk HPV negative by in situ hybridization.

\textbf{Type 6}: Descriptive \\

\bottomrule
\end{tabular}
}
\caption{Insight details of Case 4 in MedInsightBench (1).}
\label{data_example_2}
\end{table*}

\begin{table*}[htbp!]
\centering
\resizebox{\textwidth}{!}{
\begin{tabular}{p{\textwidth}}
\toprule
\textbf{Insights Details}\\
\midrule
\textbf{Question 7}: Does the pathology report indicate both the number of lymph nodes involved by metastatic carcinoma and the total number examined, along with the presence or absence of extracapsular extension?

\textbf{Insight 7}: Single lymph node metastasis (1/44) without extracapsular extension suggests intermediate recurrence risk; consider adjuvant therapy based on multidisciplinary discussion.

\textbf{Type 7}: Predictive \\
\midrule
\textbf{Question 8}: Does the pathology report confirm that all intraoperative frozen section consultations for mucosal margins were benign, indicating adequate surgical clearance?

\textbf{Insight 8}: All intraoperative frozen section consultations for mucosal margins were benign, confirming adequate surgical clearance.

\textbf{Type 8}: Evaluative \\
\midrule
\textbf{Question 9}: What are the specific benign findings and lymph node levels documented as negative for tumor in the report?

\textbf{Insight 9}: Additional benign findings include minor salivary gland tissue and multiple lymph node levels negative for tumor.

\textbf{Type 9}: Descriptive \\
\midrule
\textbf{Question 10}: What is the HPV status of the oral cavity squamous cell carcinoma as determined by testing documented in the report?

\textbf{Insight 10}: HPV-negative status in an oral cavity squamous cell carcinoma suggests non-HPV driven etiology; consider additional molecular profiling for treatment guidance.

\textbf{Type 10}: Exploratory \\

\bottomrule
\end{tabular}
}
\caption{Insight details of Case 4 in MedInsightBench (2).}
\label{data_example_3}
\end{table*}

\clearpage
\onecolumn
\section{Prompts}
\label{appendix:prompt}

\begin{figure*}[htbp]
\begin{promptbox}{Prompt for the Pathological Report Verification.}
Given the following cancer pathology report text:
<report>{report_text}</report>

Instructions:
* Analyze the given cancer pathology report text (OCR output from a PDF). Perform these checks:
    Critical checks (must pass for the report to be usable for automated data-insight extraction):
    1. Final Diagnosis / Impression present and unambiguous (parsable diagnostic phrase).
    2. Tumor size(s) present with numeric value(s) and units (e.g., "2.3 cm", "10 mm") or explicit statement that size not applicable.
    3. Tumor grade or stage info present when relevant to the specimen type (or explicit "not applicable").
    4. Lymph node status present and parsable (e.g., "0/12 nodes", "3/5 positive").
    5. Margin status present and parsable (clear: positive/negative/closest margin and measurement if given).
    6. Specimen/site and laterality clearly stated (e.g., "left lung, lower lobe").
    7. Text quality/linkability: OCR not heavily garbled (no pervasive garbage characters), and at least one linking identifier is present (accession number, specimen ID, slide ID) OR the report contains fully parsable structured key-values that allow unambiguous extraction.
    Additional helpful checks (not strictly required but increase usability):
    8. IHC / molecular results present and include marker names with interpretation (e.g., "ER: positive 90\%") if performed.
    9. Clinical history/indication present (useful for context).
    10. No internal contradictions (e.g., both "benign" and "invasive carcinoma" without explanation).
    11. Negation correctly captured for critical phrases (e.g., "no lymphovascular invasion", "negative for malignancy").

* Decision rule:
    - If ALL Critical checks (1 to 7) pass, output 1. Otherwise, output 0.
    - NOTE: If critical checks pass but any Additional check fails, still output 1 but mention the missing helpful items in the reason.

* Output format requirements (strict):
    - Your decision must be strictly enclosed in `<decision></decision>` tags and be either `1` or `0`.
    - Give your reason inside `<reason></reason>`. The reason must be concise (max ~120 words), and must include:
        - which critical checks passed/failed (brief labels, e.g., "C1:PASS; C2:FAIL"),  
        - the top 1 to 2 specific problems found (if any), and
        - a short recommended action (one of: "re-OCR", "manual pathologist review", "link accession IDs", "proceed with spot-checks").
    - Your final reply must contain only these two tags and nothing else.
    
Refer to the example responses below.
Example (acceptable):
<decision>1</decision>
<reason>PASS. Critical checks: C1,C2,C3,C4,C5,C6,C7 PASS. IHC missing (A8). Text parsable with accession present. Recommend proceeding with spot-checks and include IHC curation if available.</reason>

Example (unacceptable):
<decision>0</decision>
<reason>FAIL. Critical checks failed: C2 (tumor size missing), C7 (OCR garbling: many non-printable chars). Recommend re-running OCR with an alternate engine and manual pathologist review for affected samples.</reason>
\end{promptbox}
\end{figure*}

\clearpage
\FloatBarrier
\begin{figure*}[htbp]
\begin{promptbox}{Prompt for the Insights Generation.}
Given the following cancer pathology report text:
<report>{report_text}</report>

Given the following report evidence:
<evidence>{evidence_text}</evidence>

Instructions:
* You will analyze the given cancer pathology report (OCR output that has passed prior usability checks) and evidence. Then extract ALL pathology data insights present in the report. The number of insights may vary by report; list every distinct insight you can infer from the text.

* Insight categories (choose one per insight):
    - Descriptive: factual summaries of what the report states (e.g., specimen type, tumor size, node count, IHC results).
    - Diagnostic: statements that identify disease or etiology (e.g., "invasive ductal carcinoma", "metastatic adenocarcinoma").
    - Predictive: findings that imply future outcomes or risks (e.g., "high grade and lymphovascular invasion -> increased recurrence risk").
    - Prescriptive: specific, actionable recommendations based on findings (e.g., "recommend ER/PR testing", "suggest sentinel node biopsy").
    - Evaluative: judgements about prior interventions or response (e.g., "treatment effect present", "no residual tumor after therapy").
    - Exploratory: unexpected patterns or hypotheses worth further investigation (e.g., "discordant IHC vs morphology, consider molecular testing").

* For each insight you output, include these fields (concise, machine-parseable text inside the tag):
    - Type: one of the six categories above.
    - Insight: a concise 1 to 3 sentence paragraph that combines the observation (summary) and any actionable recommendation. If no recommendation, end with "Recommendation: none".
    - Evidence: brief quoted text or paraphrase from the report that supports the insight (include enough context to locate it).
    - Confidence: a numeric estimate from 0.0 to 1.0 reflecting how directly the report supports the insight.

* Output rules (strict):
    - Each insight must be emitted as a separate `<insight>...</insight>` block.
    - Inside each `insight` block, present fields in this exact order and simple `key: value` format (no extra markup):  
      `Type: ...; Insight: ...; Evidence: "..."; Confidence: X.X`
    - Do NOT include any text outside the `insight` tags. Your entire reply must consist only of one or more `<insight>...</insight>` blocks and nothing else.
    - Produce **all** insights you can extract; do not omit findings because they seem minor.

Refer to these examples (valid outputs):

Example - Descriptive:
<insight>Type: Descriptive; Insight: The specimen is a left lower lobectomy containing a 2.3 cm invasive adenocarcinoma; Recommendation: none. Evidence: "LEFT LOWER LOBECTOMY...invasive adenocarcinoma 2.3 cm"; Confidence: 0.95</insight>

Example - Predictive (combined):
<insight>Type: Predictive; Insight: High-grade morphology with identified lymphovascular invasion suggests increased recurrence risk; Recommendation: consider close surveillance and discuss adjuvant therapy options. Evidence: "high grade" and "lymphovascular invasion identified"; Confidence: 0.80</insight>

Example - Prescriptive (combined):
<insight>Type: Prescriptive; Insight: ER/PR status not reported while invasive carcinoma is present, so hormone-receptor testing is needed; Recommendation: order ER/PR IHC. Evidence: "ER/PR not reported; invasive carcinoma described"; Confidence: 0.70</insight>
\end{promptbox}
\end{figure*}

\clearpage
\FloatBarrier
\begin{figure*}[htbp!]
\begin{promptbox}{Prompt for the Questions Generation.}
Given the following insight type:
<type>{type}</type>

Given the following insight text:
<insight>{insight}</insight>

Given the following Evidence in the cancer pathology report text:
<evidence>{evidence}</evidence>

Instructions:
* You will be given information on a single pathology insight (insight type, insight text, and corresponding Evidence excerpt from the cancer pathology report).
* Task: produce **one** clear, concise question (ending with a question mark) that - if answered by inspecting the original pathology report - would enable an analyst to derive the given insight.
* Constraints for the generated question:
    - It **must end with a single question mark**.
    - **Do not** include any verbatim text or specific phrases from the `Evidence` field (no quoting or restating report fragments).
    - **Do not** perform analysis or give extraction rules inside the question - the question should ask *what to check* or *what confirmation is needed*, not how to compute it.
    - Prefer a single sentence; be specific enough to guide an analyst but keep wording generic (refer to "report fields", "measurements", "descriptors", etc., rather than quoting report content).
    - The question should be relevant to the insight's `Type`: Descriptive, Diagnostic, Predictive, Prescriptive, Evaluative, Exploratory.
* Output format (strict):
    - Return exactly one `<question>...</question>` tag containing only the question text (nothing else).

Example output (acceptable):
<question>...</question>
\end{promptbox}
\end{figure*}

\begin{figure*}[htbp!]
\begin{promptbox}{Prompt for the Goal Generation.}
Given the following question list:
<question_list>{question_list}</question_list>

Instructions:
* You will be given a list of concise, researchable questions derived from pathology-report insights.
* Task: analyze the question list and synthesize them into a single, integrated Goal statement that orients image-analysis and downstream research efforts.

* What the Goal must do:
    - Capture the shared analytic direction and primary objectives implied by the question set (what analysts should aim to discover or correlate in pathology images).
    - Be actionable at a high level (indicate the types of analyses or correlations to prioritize) but avoid implementation details, extraction rules, or step-by-step methods.
    - Balance scope: neither overly broad nor overly detailed - enough to guide design of image-analysis workflows and hypothesis generation.
    - Reflect clinical relevance (e.g., link morphology to outcome/markers, flag ambiguous cases for review) and encourage validation/uncertainty handling, without prescribing exact thresholds.

* Constraints:
    - Produce a single paragraph, 1 to 2 sentences long (preferably 15 to 40 words).
    - Do NOT restate or quote the input questions; synthesize their themes instead.
    - Do NOT include bullets, lists, or extra commentary.
    - The output must be strictly enclosed in a single `<goal></goal>` tag and contain only that tag and the Goal text.

Example output (acceptable):
<goal>...</goal>
\end{promptbox}
\end{figure*}

\clearpage
\FloatBarrier
\begin{figure*}[htbp!]
\begin{promptbox}{Prompt for the Image Summarization Module in Visual Root Finder.}
You are given a single pathology image of a cancerous tissue (H&E slide), and your task is to produce a concise, clinically useful summary describing what is seen. 
Do not invent clinical history or definitive diagnoses beyond what the image supports - state uncertainty where appropriate.

Output guidance (high-level, not rigid formatting):
* A short summary (brief - about 1-3 sentences) describing the main histologic features visible at low magnification (e.g., staining, overall architecture, areas of increased cellularity, gland formation, necrosis, infiltration of surrounding tissue).
* A short list of 3-6 keywords highlighting the most important features. Each keyword must be enclosed within <keyword></keyword> tags.
* One brief recommendation of next steps for diagnostic confirmation (e.g., examine higher-power fields, perform immunohistochemistry panels, correlate with clinical data).
\end{promptbox}
\end{figure*}

\begin{figure*}[htbp!]
\begin{promptbox}{Prompt for the Root Question Generator in Visual Root Finder.}
Given the following context:
<context>{context}</context>

Given the following goal:
<goal>{goal}</goal>

Given the cancer pathology image.

Given the summary of the image:
<summary>{image_summary}</summary>

Given the searching results of the image summary:
<search_results>{search_results}</search_results>

Instructions:
* Write a list of questions to be solved by your cancer pathology team to analyze the provided cancer histopathology images and reach the stated goal.
* Focus questions on image-derived evidence (slide-level labels, region/patch-level features, cellular and tissue morphology, tumor microenvironment, staining characteristics, magnification/scale, annotation masks) and any linked metadata (diagnosis, clinical outcomes, molecular markers, patient demographics).
* Explore diverse aspects of the image data and metadata, and ask questions that are directly relevant to the goal.
* To better understand and analyze the cancer pathology image, you can refer to the given summary of the image and the search results.
* You must ask the right questions to surface anything interesting in the pathology images (morphological trends, spatial patterns, rare anomalies, artifacts, staining variability, segmentation/annotation issues, correlations with outcomes, etc.).
* Make sure each question can realistically be answered using the available data schema (image tiles/patches, labels, annotations, clinical/molecular metadata, quality metrics).
* Note that the insights your team extracts will be used to generate a clinical/research report.
* Each question should be a single-part question that requires a single direct answer - end the line with exactly one '?' and avoid compound questions.
* Do not number the questions.
* You can produce at most {max_questions} questions. Stop generating after that.
* Most importantly, each question must be enclosed within <question></question> tags.

Example response:
<question>What is the tumor status and the size of the submitted lymph node from the station 7 biopsy?</question>
<question>Does the clinical history provided in the report align with the pathological diagnosis regarding the specific type of malignancy?</question>
\end{promptbox}
\end{figure*}

\clearpage
\FloatBarrier
\begin{figure*}[htbp!]
\begin{promptbox}{Prompt for the Image Analysis Tool in Analytical Insight Agent.}
### Instruction:
Given the following Question:
<question>{question}</question>

Given: one cancer pathology image (the input image) to be inspected to answer the question.

Task (what to do):
* Analyze the image with the Question above as the analytic objective. Your job is to extract the **key image-derived information** that directly relates to answering the Question, describe the visual evidence, identify the image regions to inspect, note ambiguities or limitations, and recommend next steps (additional images, stains, metadata, or human review) required to confidently answer the Question.

Required content to produce (use these exact field names and order inside the output tag):
1. `KeyImageFindings:` 2-6 concise short sentences describing the essential visual features observed that relate to the Question (morphology, pattern, structures, presence/absence of features).  
2. `RegionsOfInterest:` brief textual description of where in the image the evidence appears (e.g., "upper-left field, dense invasive nests near adipose boundary") or integer pixel/bbox coordinates if available; if none, write `none`.  
3. `Measurements:` any quantitative values you can extract/estimate from the image relevant to the Question (size in mm if slide scale known, area, counts); if none, write `none`.
4. `AmbiguitiesOrLimitations:` concise notes on what prevents a definitive answer (e.g., low resolution, focal artifact, required IHC not visible, missing context).  
5. `RecommendedNextSteps:` 1-3 short actionable recommendations to resolve ambiguities (e.g., request additional WSI, perform IHC for marker X, consult pathologist).  
6. `Confidence:` numeric score between 0.0 and 1.0 (one decimal place), estimating how confidently the image evidence supports the KeyImageFindings and a direct answer to the Question.

Constraints and style:
* Keep each field concise. Use clinical/technical wording but keep sentences short (one line each preferred).
* Assume common OCR errors and slide variability; be explicit if that affects interpretability.
* Do NOT include any narrative or extra commentary outside the required fields.

Output rules (strict):
* Your final reply must contain **only** a single `<findings>...</findings>` tag and nothing else.
* Inside the `findings` tag, present the fields in the exact order and format below, separated by semicolons (`;`) -- no other punctuation structure, no newlines outside the tag:
  `AnswerableFromImage: ...; KeyImageFindings: ...; RegionsOfInterest: ...; Measurements: ...; AmbiguitiesOrLimitations: ...; RecommendedNextSteps: ...; Confidence: X.X`
* All text must be replaceable by a downstream parser (avoid extra colons or parentheses inside field contents unless necessary).

Refer to these examples (valid outputs):

Example:
<findings>KeyImageFindings: Invasive glandular clusters with prominent nucleoli and desmoplastic stroma; RegionsOfInterest: central-right field near tissue edge; Measurements: largest tumor focus approx. 2.4 mm (estimated); AmbiguitiesOrLimitations: slide scale approximate, focal crush artifact; RecommendedNextSteps: confirm tumor size on full WSI and report scale, consider correlate with IHC if marker-specific question; Confidence: 0.8</findings>
\end{promptbox}
\end{figure*}

\clearpage
\FloatBarrier
\begin{figure*}[htbp!]
\begin{promptbox}{Prompt for the Answers \& Insights Generator in Analytical Insight Agent.}
You are trying to answer a question based on the information provided. This is a cancer pathology image dataset that could potentially consist of interesting insights

Given the goal:
<goal>{goal}</goal>

Given the question:
<question>{question}</question>

Given the analysis (if has):
<analysis>{analysis}</analysis>

Given the cancer pathology image.

Instructions:
* Based on the analysis and other information provided above, and analyze the provided cancer histopathology image, please write an answer to the question enclosed with <question></question> tags.
* The answer should be a single sentence, but it should not be too high-level and should include the key details from the justification.
* Output must use HTML-like tags in this order: first the answer between <answer></answer> tags, then the justification between <justification></justification> tags, then the insight between <insight></insight> tags. Do not output any other text outside these tags.
* The justification should concisely summarize the image-derived evidence and any relevant linked metadata (e.g., morphology, cellular atypia, mitotic figures per high-power field, necrosis extent, spatial patterns, immunostain results, tumor fraction, clinical outcome) that support the answer - keep it short (1-3 sentences).
* Use only information that can be derived from the provided histopathology images and linked metadata; do not invent patient details or data.
* The entire response must be factual, precise about uncertainty (if any), and suitable for inclusion in a clinical/research report.
* The insight should be a single, non-trivial, concise, and meaningful conclusion phrased in lay terms, grounded in the question, goal, and cancer histopathology image.
* The insight should be something interesting and grounded based on the question, goal, and cancer histopathology image, something that would be interesting. 
* Refer to the following example response for the format of the answer, justification, and insight.

Example response:
<answer>This is a sample answer</answer>
<justification>This is a sample justification</justification>
<insight>This is a sample insight</insight>
\end{promptbox}
\end{figure*}

\clearpage
\FloatBarrier
\begin{figure*}[htbp!]
\begin{promptbox}{Prompt for the Follow-Up Question Generator in Follow-Up Question Composer.}
Given the following context:
<context>{context}</context>

Given the following goal:
<goal>{goal}</goal>

Given the question and answer:
<question>{question}</question>
<answer>{answer}</answer>

Given the cancer pathology image.

Given the summary of the image:
<summary>{image_summary}</summary>

Given the searching results of the image summary:
<search_results>{search_results}</search_results>

Instructions:
* Produce a list of follow-up questions to explore the provided cancer histopathology image and reach the stated goal.
* Note that we have already answered the question and have the answer; do not include a question similar to the one above. 
* Explore diverse aspects of the cancer histopathology image, and ask questions that are relevant to my goal.
* To better understand and analyze the cancer pathology image, you can refer to the given summary of the image and the search results.
* You must ask the right questions to surface anything interesting in the pathology images (morphological trends, spatial patterns, rare anomalies, artifacts, staining variability, segmentation/annotation issues, correlations with outcomes, etc.).
* Focus questions on image-derived evidence (slide-level labels, region/patch-level features, cellular and tissue morphology, tumor microenvironment, staining characteristics, magnification/scale, annotation masks) and any linked metadata (diagnosis, clinical outcomes, molecular markers, patient demographics).
* Note that the insights your team extracts will be used to generate a clinical/research report.
* Each question that you produce must be enclosed in <question></question> tags.
* Each question should be a single-part question that requires a single direct answer - end the line with exactly one '?' and avoid compound questions.
* Do not number the questions.
* You can produce at most {max_questions} questions. Stop generating after that.

Example response:
<question>What is the tumor status and the size of the submitted lymph node from the station 7 biopsy?</question>
<question>Does the clinical history provided in the report align with the pathological diagnosis regarding the specific type of malignancy?</question>
\end{promptbox}
\end{figure*}

\begin{figure*}[htbp!]
\begin{promptbox}{Prompt for the Question Selector in Follow-Up Question Composer.}
Given the information below:
<context>{context}</context>

<goal>{goal}</goal>

<prev_questions>{prev_questions_formatted}</prev_questions>

<followup_questions>{followup_questions_formatted}</followup_questions>

Instructions:
* Given a context and a goal, select one follow-up question from the above list to explore after prev_question that will help me reach my goal.
* Do not select a question similar to the previous questions above. 
* Output only the index of the question in your response inside <question_id></question_id> tag.
* The output questions ID must be 0-indexed.

Example response:
<question_id>0</question_id>
\end{promptbox}
\end{figure*}

\end{document}